\documentclass[10pt,twocolumn,letterpaper]{article}

\usepackage[pagenumbers]{iccv} 

\usepackage{enumitem}

\definecolor{iccvblue}{rgb}{0.21,0.49,0.74}
\usepackage[pagebackref,breaklinks,colorlinks,allcolors=iccvblue]{hyperref}
\usepackage{multirow}
\usepackage{pifont}  
\usepackage{xcolor}   
\usepackage[utf8]{inputenc}
\usepackage{fontawesome}
\usepackage{graphicx} 
\usepackage{makecell}

\newcommand{\cmark}{\textcolor{green}{\ding{51}}} 
\newcommand{\xmark}{\textcolor{red}{\ding{55}}}   

\newcommand{\wyc}[1]{\textcolor{black}{#1}} 

\def\paperID{3930} 
\def\confName{ICCV}
\def\confYear{2025}

\title{FiVE~\includegraphics[width=0.05\textwidth]{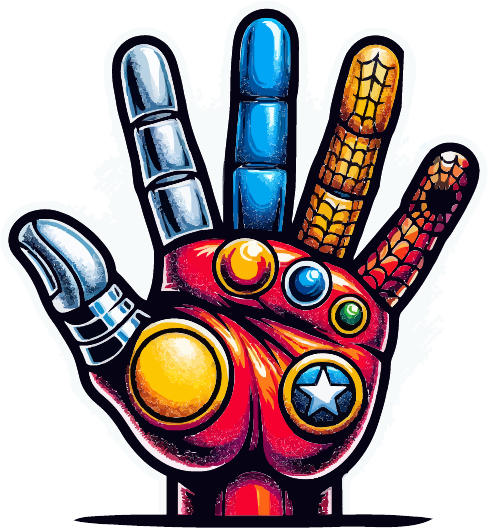}: A \underline{Fi}ne-grained \underline{V}ideo \underline{E}diting Benchmark for Evaluating \\ Emerging Diffusion and Rectified Flow Models}

\author{
Minghan Li$^{1,2}$\thanks{Equal contribution, $\dagger$ Corresponding author.}, Chenxi Xie$^{3*}$, Yichen Wu$^{4,5}$, Lei Zhang$^3$ and Mengyu Wang$^{1,2,6\dagger}$ \\
$^1$Harvard AI and Robotics Lab, Harvard University, $^2$Broad Institute, $^3$Hong Kong Polytechnic University \\
$^4$School of Engineering and Applied Sciences, Harvard University, $^5$City University of Hong Kong \\
$^6$Kempner Institute for the Study of Natural and Artificial Intelligence, Harvard University \\
{\tt\small 
mili4@meei.harvard.edu, 
chenxi.xie@connect.polyu.hk, 
yichen@seas.harvard.edu} \\
{\tt\small cslzhang@comp.polyu.edu.hk, 
mengyu\_wang@meei.harvard.edu}
}

\begin{document}

\let\oldtwocolumn\twocolumn
\renewcommand\twocolumn[1][]{%
    \oldtwocolumn[{#1}{
    \begin{center}
        \vspace{-4mm}
           \includegraphics[width=\textwidth]{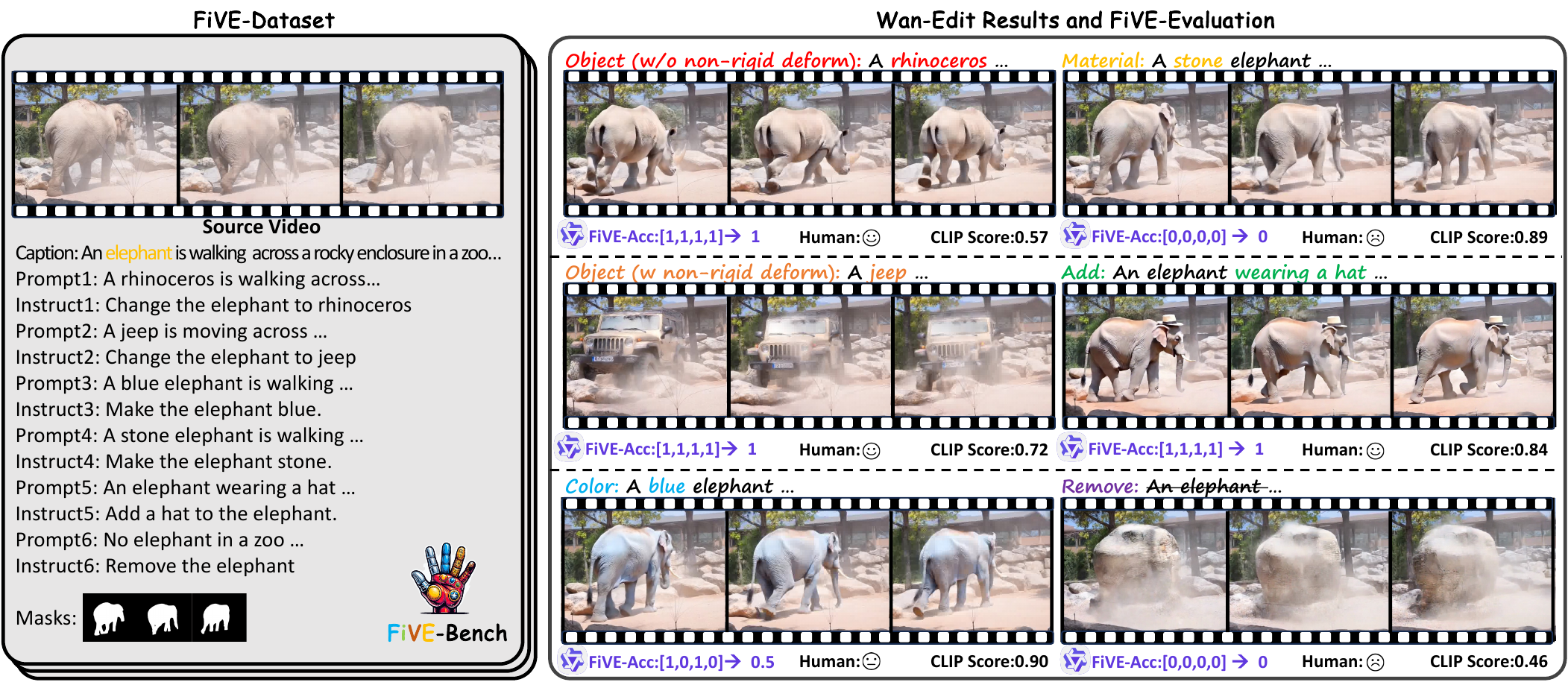}
           \vspace{-6mm}
           \captionof{figure}{{The introduced FiVE Benchmark and corresponding editing results of the proposed Wan-Edit method.}}
           \label{fig:five-bench}
           \vspace{+2mm}
    \end{center}
    }]
}

\maketitle

\begin{abstract}
Numerous text-to-video (T2V) editing methods have emerged recently, but the lack of a standardized benchmark for fair evaluation has led to inconsistent claims and an inability to assess model sensitivity to hyperparameters. Fine-grained video editing is crucial for enabling precise, object-level modifications while maintaining context and temporal consistency. To address this, we introduce \textbf{FiVE}, a \underline{Fi}ne-grained \underline{V}ideo \underline{E}diting Benchmark for evaluating emerging diffusion and rectified flow models. Our benchmark includes 74 real-world videos and 26 generated videos, featuring 6 fine-grained editing types, 420 object-level editing prompt pairs, and their corresponding masks. Additionally, we adapt the latest rectified flow (RF) T2V generation models—Pyramid-Flow~\cite{jin2024pyramidal} and Wan2.1~\cite{wan2025wan}—by introducing FlowEdit~\cite{kulikov2024flowedit}, resulting in training-free and inversion-free video editing models \textbf{Pyramid-Edit} and \textbf{Wan-Edit}. We evaluate five diffusion-based and two RF-based editing methods on our FiVE benchmark using 15 metrics, covering background preservation, text-video similarity, temporal consistency, video quality, and runtime. To further enhance object-level evaluation, we introduce \textbf{FiVE-Acc}, a novel metric leveraging Vision-Language Models (VLMs) to assess the success of fine-grained video editing. Experimental results demonstrate that RF-based editing significantly outperforms diffusion-based methods, with Wan-Edit achieving the best overall performance and exhibiting the least sensitivity to hyperparameters. More  video demo available on the anonymous website:  \url{https://sites.google.com/view/five-benchmark}
\end{abstract}    
\section{Introduction}
\label{sec:intro}

Recent advancements in text-to-image (T2I) and text-to-video (T2V) generation~\cite{rombach2022high,zeroscope,blattmann2023stable,openai2024sora,jin2024pyramidal,ma2025step,wan2025wan} have led to the emergence of numerous image and video editing methods~\cite{wu2023tune,guo2023animatediff,chen2023control,geyer2024tokenflow}, enabling users to create and modify video content with unprecedented flexibility. These methods leverage powerful generative models, such as diffusion models~\cite{ho2020denoising,song2021denoising,dhariwal2021diffusion,ho2022imagen} and \wyc{Rectified Flow}
(RF)~\cite{liu2023flow,esser2024scaling} models, to synthesize and edit videos based on textual prompts. 

However, despite significant progress in \wyc{these fields,}
the lack of a standardized benchmark for fair and comprehensive evaluation has \wyc{still} become a major bottleneck hindering further development. Currently, the video editing field has only one publicly available dataset, TGVE~\cite{wu2023TVGE} /TGVE+~\cite{singer2024videoTVGE+} (\wyc{see} Fig\wyc{.}(\ref{tab:datasets})), which includes 76 videos and 6 editing types, such as style transfer, background replacement, single-object and multi-object transformations, as well as object addition and removal. While TGVE+ aims to cover as many editing types as possible, its evaluation within each editing type is not comprehensive, particularly in supporting fine-grained video editing. Fine-grained video editing requires precise object-level modifications while preserving the overall context and temporal consistency of the video, posing higher demands on model performance.

Due to the lack of a standardized benchmark and unified evaluation framework for fine-grained video editing, existing methods~\cite{li2024vidtome,qi2023fatezero,geyer2024tokenflow,yang2025videograin} often rely on self-collected small-scale video datasets to validate their effectiveness. This approach not only makes it difficult to objectively compare the performance of different methods but also fails to comprehensively assess model robustness to hyperparameters or determine their suitability for real-world applications. Therefore, building a comprehensive and challenging benchmark for fine-grained video editing has become an urgent need to advance the field.

To address these challenges, we introduce \textbf{FiVE} (\textbf{Fi}ne-grained \textbf{V}ideo \textbf{E}diting Benchmark), a comprehensive benchmark designed to evaluate emerging diffusion and rectified flow models for fine-grained video editing. FiVE includes 74 real-world videos and 26 generated videos, covering a diverse range of scenes and editing scenarios. It features 6 fine-grained editing types, 420 object-level editing prompt pairs, and their corresponding masks, providing a rich and challenging testbed for evaluating video editing methods. Additionally, we propose FiVE-Acc, a novel evaluation metric that leverages Vision-Language Models (VLMs) to assess the successful accuracy of fine-grained object-level editing. \wyc{Specifically,} FiVE-Acc complements traditional metrics by providing a more nuanced understanding of editing quality, particularly in terms of semantic alignment, contextual preservation and motion awareness.

Recently, many image editing methods~\cite{rout2024semantic,wang2024taming,deng2024fireflow,kulikov2024flowedit} based on the latest rectified flow T2I generation model Flux~\cite{flux_github} have achieved state-of-the-art editing results and fidelity, such as RF-Inversion~\cite{rout2024semantic}, RF-Solver~\cite{wang2024taming}, and FlowEdit~\cite{kulikov2024flowedit}. However, to the best of our knowledge, there are no video editing methods based on RF T2V generation models, primarily due to the relatively slow development of T2V generation models. Fortunately, the recent release of two state-of-the-art RF-based T2V generation models—Pyramid-Flow~\cite{jin2024pyramidal} and Wan2.1~\cite{wan2025wan}—has opened new possibilities for RF video editing.

Building on the latest advancements, we adapt Pyramid-Flow~\cite{jin2024pyramidal} and Wan2.1~\cite{wan2025wan} by introducing FlowEdit~\cite{kulikov2024flowedit}. This adaptation \wyc{facilitates} training-free and inversion-free video editing models, Pyramid-Edit and Wan-Edit, which leverage the strong temporal consistency inherent in video generation models without requiring additional attention mechanisms. Specifically, Pyramid-Flow employs a multi-resolution and temporally autoregressive model architecture across time steps. To adapt to this unique design, we modified FlowEdit to support multi-resolution processing, enabling seamless integration with the Pyramid-Flow framework, \wyc{leading to} Pyramid-Edit. On the other hand, Wan2.1 utilizes the DiT architecture, which unfolds temporal frames into sequences and processes them via self-attention. Since its design is fundamentally similar to image generation models, Wan2.1 can directly adopt FlowEdit without any additional modifications, giving rise to Wan-Edit.

\wyc{To show the effectiveness of the proposed two RF-based editing methods, we compare them with}
five diffusion-based editing methods on the \wyc{built} FiVE benchmark, evaluating them across 14 metrics, including background preservation, edited text-video similarity, temporal consistency, and generated video quality. Our experiments demonstrate that RF-based editing methods significantly outperform diffusion-based methods across multiple metrics. Notably, Wan-Edit achieves the best overall performance, exhibiting superior editing quality and the least sensitivity to hyperparameters. These findings highlight the potential of RF-based models for fine-grained video editing and provide valuable insights for future research in this area.

In summary, this work makes four key contributions:
\begin{itemize}
    \item We introduce \textbf{FiVE}, a comprehensive benchmark for fine-grained video editing, featuring diverse videos, editing types, and evaluation metrics.
    \item We propose \textbf{FiVE-Acc}, a novel metric leveraging VLMs to assess the success of fine-grained editing.
    \item We adapt RF-based T2V models Pyramid-Flow and Wan2.1 using FlowEdit, resulting in efficient and effective video editing methods \textbf{Pyramid-Edit} and \textbf{Wan-Edit}.
    \item \wyc{We evaluate seven video editing methods across 15 metrics, showing that RF-based editing consistently outperforms diffusion-based approaches.}
\end{itemize}

\begin{table*}[t!]
    \centering
    \caption{Statistics of existing video editing datasets and benchmarks. 
    }\label{tab:datasets}
    \vspace{-2mm}
    {\footnotesize
    \resizebox{\textwidth}{!}{
    \renewcommand{\arraystretch}{1.0}
    \setlength{\tabcolsep}{2pt} 
    \begin{tabular}{l c cccc c cccc l}
        \toprule
        \multirow{2}{*}{{Dataset}} &\multirow{2}{*}{Usage} &{Num.} &{Num.} &{Frames} &{Num.} &{Gen.} &{Src. obj.} &{Edit obj.} &{Obj.} &{Src. obj.} &\multirow{2}{*}{{Type of Edited Prompts}} \\
         & {} &Images &{Videos} &{Per Video} &{Prompts} &{Videos} &{Words} &{Words}  &{Instruct.} &{Masks} & \\
        \midrule
        VIVID-10M   & Train &672K &73.7K  &30         &10M &\xmark &\xmark &\xmark &\cmark &\cmark &{object, add, remove}  \\
        Señorita-2M & Train &0    &388.9K &33$\sim$64 &2M  &\xmark &\xmark &\xmark &\cmark &\xmark &{style, object, multiple, color, add, remove, motion} \\
        \midrule
        TGVE        &Eval &0 &76  &32 &304 &\xmark &\xmark &\xmark&\xmark &\xmark &{style, object, bg, multiple} \\
        TGVE+       &Eval &0 &76  &32 &1417 &\xmark &\xmark&\xmark &\xmark &\xmark &{style, object, bg, multiple, color, add, remove} \\
        VIVID-10M   &Eval &0 &64  &33$\sim$64    &852  &\xmark &\xmark &\xmark &\cmark &\cmark &{object, add, remove} \\
        FiVE (Ours) &Eval &0 &100 &35 $\sim$ 126 & 420 &\cmark &\cmark &\cmark &\cmark&\cmark & object (w\&w/o non-rigid), color, material, add, remove \\
        \bottomrule
    \end{tabular}
    }
    }
\end{table*}
\section{Related Work}
\label{sec:related_work}

\begin{figure*}
    \centering
    \includegraphics[width=0.97\linewidth]{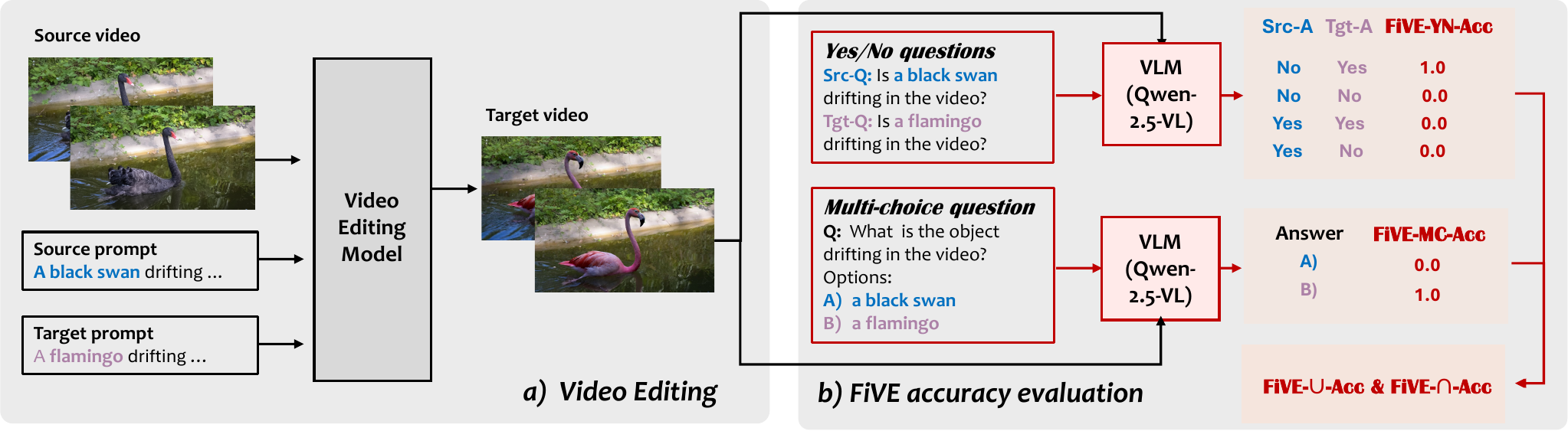}
    \vspace{-1mm}
    \caption{The VLM-based FiVE accuracy (FiVE-Acc) evaluation.}
    \label{fig:FiVE-Acc}
\end{figure*}

\subsection{Diffusion and RF Inversion}

\textbf{Diffusion inversion.} Diffusion models (DMs) drive advances in inversion techniques for better control over generated samples. DDIM~\cite{song2021denoising} accelerates sampling but struggles with fine-grained reconstructions due to nonlinearities and score estimation errors. To improve accuracy, fine-tuning and optimization-based methods~\cite{mokady2023null,dong2023prompt,wallace2023edict,wu2023latent,ju2023direct} have been proposed. While these advancements significantly enhance the fidelity and accuracy of DDIM inversion, the associated computational costs and resource demands remain substantial limiting factors. 
Although diffusion-based methods have achieved outstanding performance, recent studies~\cite{esser2024scaling,rout2024semantic,flux_github,jin2024pyramidal} suggest that Rectified Flow models (RFs) hold significant potential to surpass DMs in certain applications. Notably, the inversion and editing capabilities of RF models have been explored in image editing~\cite{rout2024semantic,kulikov2024flowedit,deng2024fireflow,wang2024taming,xie2025dnaedit}, demonstrating their efficiency and effectiveness. Building on these strengths, this work extend their potential in video editing. 

\subsection{Video Editing}
Video editing can generally be categorized into three types based on their training: training-free methods, one/few-shot finetuned methods, and massive data finetuned methods. Training-free models \cite{yang2025videograin,qi2023fatezero,li2024vidtome,ku2024anyv2v,jiang2025vace}, enable task execution without retraining. TokenFlow~\cite{geyer2024tokenflow} uses pre-trained models for efficient video synthesis, while DMT~\cite{yatim2024space} transfers motion via diffusion. Other models like Pixel2Video~\cite{ceylan2023pix2video}, Render-A-Video~\cite{yang2023rerender}, and Text2Video-zero~\cite{khachatryan2023text2video} enhance accessibility for real-time video editing. One/few-shot fine-tuned models adapt pre-trained models for video editing with minimal fine-tuning. Methods like Tune-A-Video~\cite{wu2023tune}, MotionDirector~\cite{zhao2025motiondirector}, DreamVideo~\cite{wei2024dreamvideo}, and MotionEditor~\cite{tu2024motioneditor} achieve strong results using limited annotated samples, balancing customization and efficiency for task-specific editing.

Massive data fine-tuned models undergo extensive training on large datasets for high-quality, versatile video editing~\cite{qin2024instructvid2vid,jiang2025vace,tan2025omni}. InstructVid2Vid~\cite{qin2024instructvid2vid} enables complex edits via natural language instructions, while EffiVED~\cite{zhang2024effived} refines broad datasets into high-quality subsets for efficient editing. SF-V~\cite{zhang2024sf} introduces adversarial training for video synthesis. Despite their precision and adaptability, these methods are computationally expensive, limiting practicality in resource-constrained settings.
These methods focus on learning video motion, which may compromise background preservation. Recently, VideoGrain~\cite{yang2025videograin} has emphasized multi-granularity video editing rather than fine-grained edits, aiming to retain background details while modifying target objects. This work aims to address this issue and advance the field.

Recent unified architectures~\cite{jiang2025vace,tan2025omni,ye2025unic} have also incorporated video editing into their multi-functional frameworks. VACE ~\cite{jiang2025vace} enables users to perform video generation, editing, and personalization in an integrated manner. Such architectures typically concatenate reference and target tokens as input to the model, allowing for a unified processing pipeline.

\section{FiVE Benchmark and FiVE-Acc}
\label{sec:benchmark}
\wyc{In this section, we present the FiVE benchmark, which includes the FiVE video dataset, six fine-grained editing tasks, and the proposed VLM-based FiVE-Acc metric to evaluate editing accuracy.}


\subsection{FiVE Benchmark}

\textbf{Videos in FiVE Benchmark.}~We meticulously curate 74 real-world videos suitable for fine-grained editing from the DAVIS~\cite{perazzi2016davis} dataset. Consecutive frames are extracted at every 8-frame interval, and the GPT-4o~\cite{hurst2024gpt} is employed to generate formatted captions for these videos. These captions encompass details pertaining to object category, action, background, and camera movement. Additionally, during the annotation process, we document instances of object deformation (such as limb movements in humans or animals), which aids in differentiating the complexity of editing tasks during evaluation.
To further enhance the diversity of our benchmark, we generate 26 highly realistic synthetic videos using the T2V model~\cite{wan2025wan}. These synthetic videos not only expand the range of video categories but also enable a comparative analysis of editing performance across real and synthetic videos. \wyc{More details on the selection of text prompts for video generation are in the Appendix.}

\textbf{Six Fine-grained Video Editing Tasks.}
FiVE comprises six fine-grained video editing tasks, totaling \textbf{420} high-quality editing prompt pairs. It primarily includes four object-targeted editing tasks of increasing complexity: \textit{color alteration}, \textit{material modification}, \textit{object substitution without non-rigid deformation}, and \textit{object substitution with non-rigid deformation}. For each task, we use GPT-4o~\cite{hurst2024gpt} to generate four source-target prompt pairs per video in the benchmark, resulting in 400 high-quality editing pairs. To further enhance the diversity of editing types within the benchmark, we select 10 videos each for the \textit{add} and \textit{remove} tasks and design corresponding source-target prompt pairs.
Additionally, we explicitly provide both object-related source words and edited words in the source and edited text prompts, respectively, denoted as `src. obj. words' and `edit obj. words' in Table \ref{tab:datasets}. To accommodate different editing methods, we use GPT-4o~\cite{hurst2024gpt} to generate instruction prompts (`Obj. Instr.' in Table~\ref{tab:datasets}) for each source-target pair, ensuring compatibility with models like InstructPix2Pix~\cite{brooks2023instructpix2pix}. For each video, we employ SAM2~\cite{ravi2024sam2segmentimages,li2024univs} to generate masks of the edited regions, enabling the evaluation of background preservation metrics. Additional details can be found in the Appendix.

\subsection{FiVE-Acc Evaluation}
To evaluate the accuracy of fine-grained video editing, we propose FiVE accuracy evaluation, a framework assessing how precisely video editing models modify the target object. Fig. \ref{fig:FiVE-Acc} presents the FiVE pipeline, comprising video editing and accuracy evaluation.

\textbf{Video editing.} Given a source video, a video editing model is prompted with a source prompt, describing the initial scene, and a target prompt, specifying the desired modification. The model generates a target video, where the intended transformation is applied.

\textbf{FiVE accuracy evaluation.}
To quantitatively evaluate the fidelity of video editing, we employ a Vision-Language Model (VLM) (\textit{e.g.,} Qwen-2.5-VL) to analyze the edited video. The evaluation includes two types of questions:
\begin{itemize}
    \item Yes/No Questions: The model is asked whether the source and target objects are present in the target video. Editing succeeds only if the model answers `No' for the source object and `Yes' for the target, leading to the calculation of \textit{FiVE-YN-Acc}. The source object question ensures the model both removes the source and adds the target, refining accuracy evaluation.
    \item Multi-choice Question: It evaluates whether the VLM recognizes the target (e.g., flamingo in Fig. \ref{fig:FiVE-Acc}) or source object (e.g., black swan in Fig. \ref{fig:FiVE-Acc}) in the edited video. The model selects between the two objects to compute \textit{FiVE-MC-Acc}, measuring its recognition accuracy.
\end{itemize}

Based on the two obtained accuracies, we further calculate the union and intersection accuracies to derive \textit{FiVE-$\cup$-Acc} and \textit{FiVE-$\cap$-Acc}. These offer a more detailed evaluation: the former reflects the model's overall editing success accuracy, while the latter highlights the model's high-quality editing success accuracy. In summary, FiVE-Acc combines the accuracies of four components, using the VLM's recognition ability to evaluate the alignment between the edited object and its real data distribution. To the best of our knowledge, we are the first to introduce accuracy evaluation in fine-grained video editing.
\section{Methods}
\label{sec:method}

\begin{figure*}[t]
    \vspace{-4mm}
    \centering
    \begin{subfigure}[b]{0.48\linewidth}
        \centering
        \includegraphics[width=0.99\linewidth]{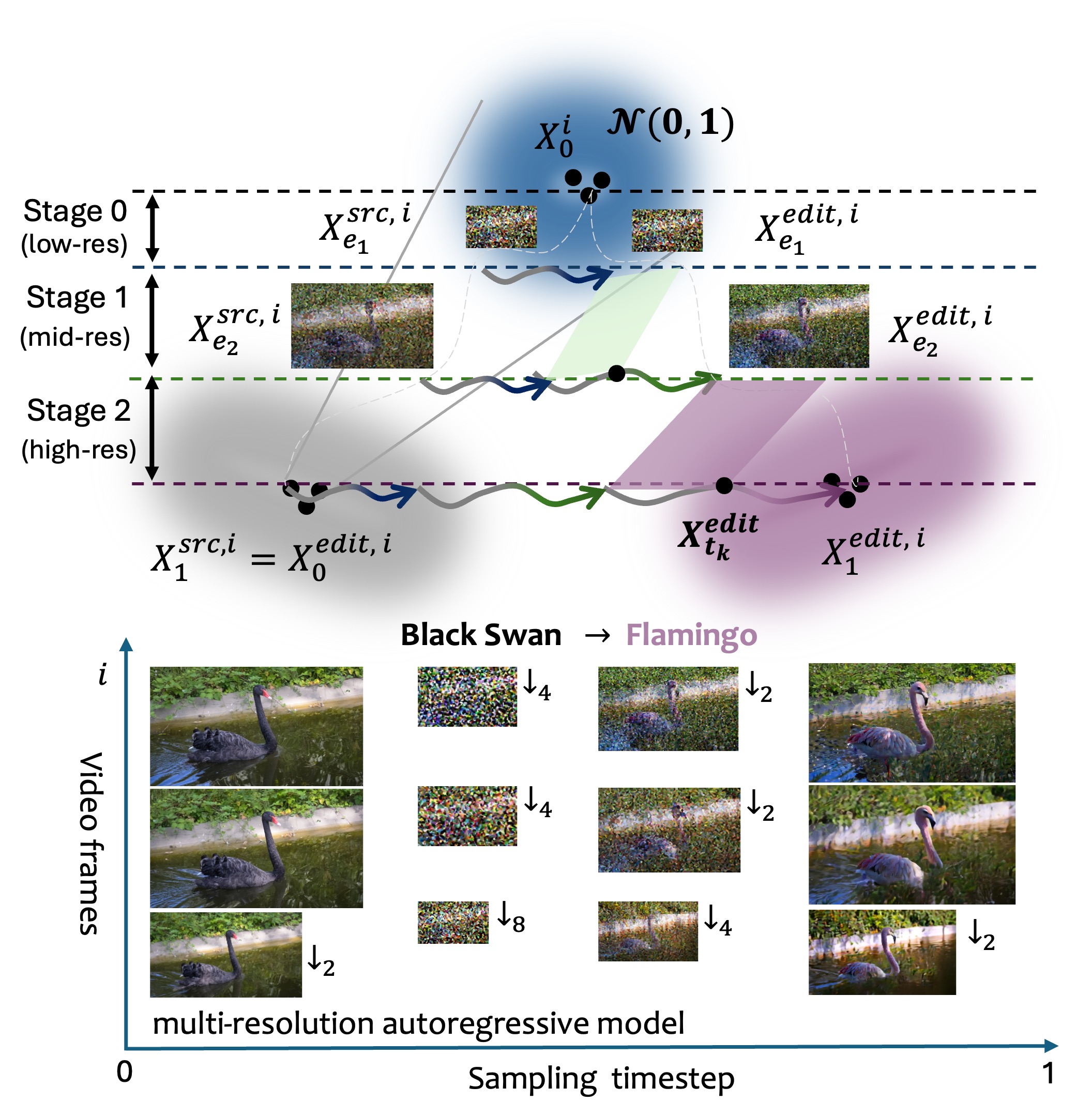}
        \vspace{-1mm}
        \caption{Pyramid-Edit}
        \label{fig:pyramid-edit}
    \end{subfigure}
    \hfill
    \begin{subfigure}[b]{0.48\linewidth}
        \centering
        \includegraphics[width=0.99\linewidth]{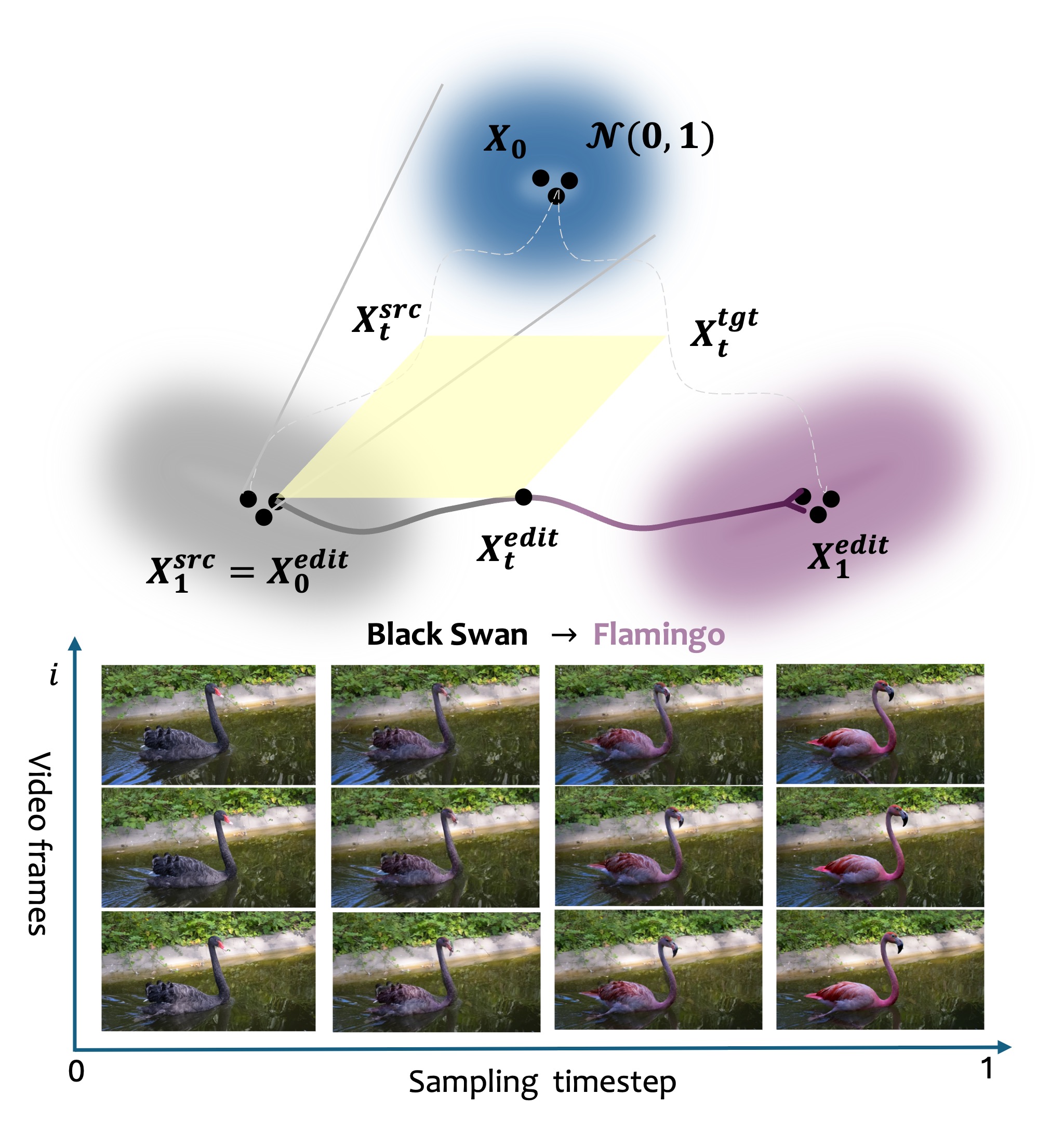}
        \vspace{-1mm}
        \caption{Wan-Edit}
        \label{fig:wan-edit}
    \end{subfigure}
    \vspace{-2mm}
    \caption{Two RF-based inversion-free video editing models: Pyramid-Edit and Wan-Edit. Pyramid-Edit is a multi-resolution temporal autoregressive architecture, while Wan-Edit (DiT architecture) treats temporal frames as a sequence and processing them simultaneously.}
    \label{fig:rf-edit}
\end{figure*}

In this section, we first provide a brief review of rectified flow (RF) and the RF-based image editing method, FlowEdit. We then introduce two RF-based video editing methods, PyramidEdit and WanEdit.

\subsection{Preliminary}
\textbf{Rectified Flow (RF) Models~\cite{lipman2023flow,liu2023flow} .}
Let $q_0$ denote the source distribution (i.e., standard Gaussian $\mathcal{N}(0,I)$), $p_0$ refer to the target distribution (i.e., the distribution over images), and $v_t(\cdot)$ is the time-varying vector field. RF models~\cite{lipman2023flow,liu2023flow} aim to transform $q_0$ progressively into $p_0$ by solving the following ordinary differential equation (ODE):
\begin{equation}
\label{eqns:RFs}
    dX_t =v_t(X_t)dt, ~~ t\in[0,1].
\end{equation}
Here, the vector field $v_t(X_t)= f(X_t, c, 1-t; \varphi)$, where $f(\cdot)$ is a neural network parameterized by $\varphi$ and $c$ is the guided text. Given the ODE in Eq.~(\ref{eqns:RFs}), we start with $X_0 \sim q_0$ from the source distribution and integrate over $t: 0\rightarrow 1$ to reach the endpoint $X_1 \sim p_0$, which is sampled from the target distribution. the forward process of RF assumes the linear path between the two states: $X_t = (1-t)X_0 + tX_1$, then we can derive the vector field of $v_t(\cdot)$: 
$u_t(X_t) = X_1-X_0$.
Then, $v_t(X_t)$ is used to approximate the network $f(X_t, c, t;\varphi)$ by minimizing the loss function
    $\mathcal{L}_\varphi= \mathbb{E}_{t\sim\mathcal{U}[0,1], Y_t}[ \|v_t(X_t)-f(X_t,c, t;\varphi) \|_2^2].$
RF ensures that their sampling paths are relatively straight, enabling the use of a small number of discretization steps.

\textbf{RF-based Image Editing FlowEdit~\cite{kulikov2024flowedit}.}
Image editing involves taking a source image $X_1^{src}$ and a source text prompt $c^{src}$, and allowing the user to provide a target text prompt $c^{tgt}$ for targeted fine-grained editing to ultimately yield the edited image $X_1^{tgt}$. For example, the source text prompt might be `A black swan is swimming,' while the target text prompt could be `A flamingo is swimming.' To simplify notation, we denote the vector fields corresponding to the source and target text prompt as $v_t(X_t^{src})$ and $v_t(X_t^{tgt})$, respectively.
Their sampling process in RF follows:
$
    \label{eq:rf_src_tgt}
    dX_t^{src} =v_t(X_t^{src})dt,\ dX_t^{tgt} =v_t(X_t^{tgt})dt.
$

FlowEdit operates inversion-free, mapping source modes to the nearest target modes, as shown in the right of Fig. \ref{fig:rf-edit}. It assumes target images follow straight trajectories, allowing latent interpolation between image and noise: $X_t^{src} = (1-t) X_0 + t X_1^{src}$, where $X_0$ is randomly sampled from Gaussian noise. Thus, the nearest target mode $X_t^{edit}$ in timestep $t$ is derived by subtracting the source image from the difference between the source and target latents:
\begin{equation}
    \label{eq:edit}
    X_t^{edit} = X_1^{src} + X_t^{tgt} - X_t^{src}.
\end{equation}
Note that $X_t^{edit}$ remains within the clean image latent space without incorporating any noise, following:
$X_t^{edit} = X_1^{src}, $ if $ t  \rightarrow 0$; $X_t^{edit} = X_t^{tgt}, $ if $ t  \rightarrow 1$. A more straightforward explanation is: $ X_t^{edit} $ evolves within the target distribution, continuously transitioning through the clean latent space across all timesteps, ({\it i.e.} $ X_t^{edit} $ gradually transforms from a black swan to a flamingo as the timestep progresses in Fig. \ref{fig:rf-edit}). Therefore, we can obtain $X_t^{tgt}$ simply by solving for $ X_t^{edit},\ t \rightarrow 1 $. By substituting Eq. (\ref{eq:edit}), we derive: $X_t^{tgt}=X_t^{edit}\!+\!X_t^{src}\!-\! X_1^{src}$. Since $X_t^{edit}$ is unknown at the current timestep, we approximate it using $X_{t-1}^{edit}$, yielding: 
\begin{equation}
    \label{eq:tgt}
    X_t^{tgt} \approx \hat{X}_t^{tgt}=X_{t-1}^{edit}\!+\!X_t^{src}\!-\! X_1^{src}.
\end{equation}
Finally, the editing process can be formulated as:
\begin{align}
    \label{eq: flowedit}
    X_t^{edit} &= X_{t-1}^{edit} + dX_t^{tgt} - dX_t^{src} \nonumber \\
               &\approx X_{t-1}^{edit} + \mathbb{E}[ v_t(\hat{X}_t^{tgt})- v_t(X_t^{src})|X_1^{src}]dt \nonumber \\
               & \triangleq X_{t-1}^{edit} + \mathbb{E}[ v^\Delta_t(\hat{X}_t^{tgt}, X_t^{src})|X_1^{src}]dt
\end{align}
To obtain the edited image, $ X_t^{edit} $ is iteratively updated from $ \epsilon $ to 1, where $ \epsilon \in [0, 1) $ represents skipped timesteps. 
The edited latent $ X_\epsilon^{edit} $ is initialized as $ X_1^{src} $. The source latent $ X_t^{src} $ is obtained via interpolation between random noise and the source image, while $ X_t^{tgt} $ is computed using Eq. (\ref{eq:tgt}). Finally, $ X_t^{edit} $ is updated following Eq. (\ref{eq: flowedit}).

\subsection{RF-based Video Editing}
This section introduces Pyramid-Edit, followed by a discussion of Wan-Edit, highlighting key features and differences.

\subsubsection{Pyramid-Edit}
\textbf{Pyramid-Flow}~\cite{jin2024pyramidal} employs a multi-resolution scheme across timesteps and a temporally autoregressive architecture to effectively handle the spatial-temporal complexity in video generative modeling.~It decomposes the flow into segments over $K$ timestep windows within the interval [0,1].~Each window interpolates between successive resolutions, reducing redundant computations in the earlier steps.~For example, in the sampling process, within the $k$-th window~$[s_k,e_k]$, it defines the rescaled timestep {\small$t_k\!=\!(t-s_k)/(e_k-s_k)$} and the corresponding flow is 
\begin{equation}
\begin{split}
    X_{t_k}   &= (1-t_k)\bar{\mathbf{x}}_{s_k}+ t_k \bar{\mathbf{x}}_{e_k}, \\
    \bar{\mathbf{x}}_{s_k} & = (1\!-\!s_k) X_0 + s_k Up(Down({X_1,2^{k+1}})), \\
    \bar{\mathbf{x}}_{e_k} & = (1\!-\!e_k) X_0 + e_k Down({X_1,2^{k}}), 
\end{split}
\end{equation}
where {\small $Down(\cdot,\cdot)$} and {\small $Up(\cdot)$} represent the down-sampling and up-sampling, respectively.~Meanwhile, the vector field in the $k$-th window is defined as {\small $v_{t_k}(X_{t_k})=\bar{\mathbf{x}}_{e_k}-\bar{\mathbf{x}}_{s_k}$}. To maintain continuity in the probability path within the pyramid structure, corrective noise is added at transition points between stages, as follows:
{\small
\begin{equation}
    \label{eq:upsample}
     X_{s_{k-1}} = \frac{s_{k-1}}{e_k}Up(X_{e_k}) \!+\! \alpha \mathbf{n}', s.t.~\mathbf{n}_{k-1}\!\sim\! \mathcal{N}(0,\mathbf{\Sigma}'),
\end{equation}
}
\!\!where the rescaling coefficient $s_{k-1}/e_k$ aligns the means of these distributions, the corrective noise $\mathbf{n}_{k-1}'$ weighted by $\alpha$ aligns their covariance matrices, and the covariance matrix $\mathbf{\Sigma}'$ is related to the upsampling function. 

\textbf{Pyramid-Edit.} The autoregressive architecture allows Pyramid-Edit to process frames sequentially, where previous frames serve as conditional information. However, the multi-resolution scheme across timesteps hinders direct use of FlowEdit. To address this, Pyramid-Edit integrates FlowEdit (Eq. (\ref{eq: flowedit})) into each window as follows:
\begin{align}
    \label{eq:pyramidedit}
    X_{t_k}^{edit} &= X_{t_k-1}^{edit} + \mathbb{E}[v^\Delta_{t_k}(\hat{X}_{t_k}^{tgt}, X_{t_k}^{src})|\bar{\mathbf{x}}_{e_k}^{src}]dt,
\end{align}
where $\hat{X}_{t_k}^{tgt} = X_{t_k-1}^{edit}\!+\!X_{t_k}^{src}\!-\! \bar{\mathbf{x}}_{e_k}^{src}$, and the target distribution is the end point of each window {\small$\bar{\mathbf{x}}_{e_k}^{src}$}, instead of the clean latent {\small $X_1^{src}$} in FlowEdit.

The workflow proceeds as illustrated in Fig. \ref{fig:rf-edit} (a) and is structured as follows: starting from noise and the lowest-resolution latent, {\small $X_{t_0}^{edit}$} is initialized as {\small $\bar{\mathbf{x}}_{e_0}^{src}$}, Eq.~(\ref{eq:pyramidedit}) is used to reconstruct and edit low-frequency information. At the end of the first stage, the reconstructed latent and the edited latent are upsampled to a higher resolution and passed through corrective noise as described in Eq.~(\ref{eq:upsample}) to obtain the starting point for the next stage, {\small $X_{s_1}^{edit}$} and {\small $X_{s_1}^{src}$}. 
However, FlowEdit operates within the target distribution, meaning we require the end point for this stage, {\small $\displaystyle X_{e_1}^{\text{edit}}$ } and { \small $\displaystyle X_{e_1}^{src}$}. The latter, {\small $\displaystyle X_{e_1}^{src}$}, can be obtained through interpolation between {\small $\displaystyle X_0$} and {\small $\displaystyle X_1^{src}$}, \textit{i.e.} ({\small $\displaystyle X_{e_1}^{src} = \bar{\mathbf{x}}_{e_1}^{src}$}). To estimate the edited endpoint {\small $\displaystyle X_{e_1}^{\text{edit}}$}, we first compute the difference between the start and end points of the source prompt in this stage. This difference is then added to the edited start latent variable, providing an estimate of {\small $\displaystyle X_{e_1}^{\text{edit}}$}:
\begin{align}
    \bar{\mathbf{x}}_{e_1}^{src} &= X_{s_1}^{src} + (\bar{\mathbf{x}}_{e_1}^{\text{src}} - X_{s_1}^{src}), \\
    X_{e_1}^{\text{edit}} &= X_{s_1}^{\text{edit}} + (\bar{\mathbf{x}}_{e_1}^{\text{src}} - X_{s_1}^{src}). 
\end{align}
\noindent 
This step ensures that the edited and reconstructed latents remain aligned with the source image distribution while incorporating the desired modifications, preserving the integrity of the generated features. This is key to the success of FlowEdit under the pyramid structure.

Another key point is that, starting from the second frame, Pyramid-Edit respectively incorporates the source and edited historical information from previous frames (e.g., the {\small $(i\!\!-\!\!1)$}-th frame) as a conditioning factor within its noise-estimation network. This allows the model to leverage temporal dependencies from both the source and edited frames, enhancing the temporal consistency. The source and edited sampling processes can be simplified as follows:
{\scriptsize
\begin{align}
     \underbrace{...\rightarrow Down(X_1^{src,i\!-\!2},2^{k+1}) \rightarrow Down(X_1^{src, i\!-\!1}, 2^k)}_{\text{Source History condition}} \rightarrow  X_{t_k}^{src, i}, \nonumber\\
     \underbrace{...\rightarrow Down(X_1^{edit,i\!-\!2},2^{k+1}) \rightarrow Down(X_1^{edit, i\!-\!1}, 2^k)}_{\text{\quad Edited History condition}} \rightarrow  X_{t_k}^{edit, i}. \nonumber
\end{align}
}

\noindent For generating subsequent frames, simply introduce the superscripts $ i $ as indices for the historical latents in Eq. (\ref{eq:pyramidedit}).

\begin{table*}[ht]
    \centering
    \caption{Comparison of diffusion- and flow-based video editing methods on our proposed FiVE benchmark.$^*$ and $\dagger$ denote methods that require optimization and depth/segmentation maps, respectively.}
    \vspace{-2mm}
    \resizebox{0.99\textwidth}{!}{
    \renewcommand{\arraystretch}{1.2}
    \setlength{\tabcolsep}{2.2pt} 
    \begin{tabular}{cl|c|cccc|cc|c|c|c}
        \Xhline{1pt}
        \multicolumn{2}{c|}{\multirow{2}{*}{\textbf{Methods}}} & \textbf{{Structure}} &\multicolumn{4}{c|}{\textbf{Background Preservation}} &\multicolumn{2}{c|}{\textbf{Text Alignment}} &\multicolumn{1}{c|}{\textbf{IQA}} &\multicolumn{1}{c|}{\textbf{Motion}} &\textbf{Time (s)} \\
        & &Dist.$_{\times 10^3}$$\downarrow$ &PSNR$\uparrow$ &LPIPS$_{\times 10^3}$$\downarrow$ &MSE$_{\times 10^4}$$\downarrow$ &SSIM$_{\times 10^2}$$\uparrow$ &CLIPS.$\uparrow$ &CLIPS.$_{\text{edit}}\uparrow$ &NIQE$\downarrow$ & Fidelity S.$_{\times 10^2}$$\uparrow$ &Per Frame$\downarrow$ \\
        \Xhline{1pt} 
        &Source Videos &0.00 & $\infty$ &0.00 &0.00 &100.00 &24.59 & 19.87 &6.33 &93.76 &-\\
        \hline
        \multirow{5}{*}{DMs} 
        &TokenFlow~\cite{geyer2024tokenflow}  &35.62 & 19.06 & 263.61 & 138.65 & 72.51 & 26.46 & 21.15 &\textbf{4.01} &{89.00} &8.04 \\
        &DMT$^*$~\cite{yatim2024space}  &85.95 & 14.71 & 404.60 & 372.78 & 51.64 & 26.66 & \textbf{21.44} &5.24 &82.30 &25.98 \\
        &VidToMe~\cite{li2024vidtome}  &22.37 & 21.15 & 263.91 & 88.75 & 70.69 & \textbf{26.84} & 21.05 &4.68 &\textbf{90.06} &3.25 \\
        & AnyV2V~\cite{ku2024anyv2v}  &71.36 & 15.90 & 348.59 & 342.97 & 50.77 & 24.89 & 19.72 &5.04 &60.36 &6.11 \\
        &VideoGrain$^{\dagger}$~\cite{yang2025videograin} &\textbf{12.40} & \textbf{27.05} & \underline{185.21} & \textbf{25.10} & \underline{79.13} & 25.69 & 20.31 &\underline{4.08} &88.57 &27.12 \\
        \hline
        \multirow{2}{*}{\makecell{RFs \\ (Ours)}} 
        &Pyramid-Edit  &28.65 & 20.84 & 276.59 & 95.63 & 71.72 &\underline{26.82} & 20.20 &5.48 &80.59 &\textbf{1.44} \\
        &Wan-Edit      &\underline{12.53} & \underline{25.57} & \textbf{94.61} & \underline{41.84} &\textbf{82.55} & 26.39 & \underline{21.23} &6.54 &\underline{89.43}  &\underline{3.07} \\
        \Xhline{1pt}
    \end{tabular}}
    \label{tab:basellines}
\end{table*}

\begin{table}[ht]
    \centering
    \caption{Comparison of diffusion- and flow-based video editing methods on the FiVE benchmark using FiVE-Acc metrics.}
    \label{tab:basellines_five_acc}
    \vspace{-2mm}
    \footnotesize
    \renewcommand{\arraystretch}{1.2}
    \setlength{\tabcolsep}{2.3pt} 
    \begin{tabular}{l|cccc|c}
        \Xhline{1pt}
        \multicolumn{1}{l|}{Method} &{FiVE-YN} &{FiVE-MC} &{FiVE-$\cup$} &{FiVE-$\cap$} & FiVE-Acc$\uparrow$ \\
        \Xhline{1pt}
        TokenFlow~\cite{geyer2024tokenflow}  &19.36 & 35.51 & 36.68 & 18.18 & 27.43 \\
        DMT$^*$~\cite{yatim2024space}  &\underline{34.78} & \textbf{62.06} & \textbf{62.98} & \underline{33.86} & \textbf{48.42} \\
        VidToMe~\cite{li2024vidtome}   &20.03 & 33.50 & 36.20 & 17.34 & 26.77 \\
        AnyV2V~\cite{ku2024anyv2v}  &30.62 & 45.42 & 48.96 & 27.09 & 38.02 \\
        VideoGrain$^\dagger$~\cite{yang2025videograin} &30.50 & 43.97 & 44.30 & 30.17 & 37.23 \\
        \hline
        Pyramid-Edit  &33.67 & \underline{54.01} & \underline{56.36} & 31.31 & 43.84 \\
        Wan-Edit      &\textbf{41.41} & 52.53 & 55.72 & \textbf{38.22} & \underline{46.97} \\
        \Xhline{1pt}
    \end{tabular}
\end{table}

\subsubsection{Wan-Edit}

\textbf{Wan2.1}~\cite{wan2025wan} is a state-of-the-art video generative model built on the mainstream Video Diffusion DiT framework~\cite{openai2024sora}, achieving significant advancements through innovations such as a novel 3D VAE, scalable training strategies, large-scale data curation.~These contributions enable Wan2.1 to generate high-quality, temporally consistent videos with improved efficiency and scalability. However, since the technical report has not yet been publicly released, we focus on adapting its core framework for video editing tasks, leveraging its strengths in temporal modeling and high-fidelity generation.

Compared to Pyramid-Flow, Wan2.1 adopts a much simpler architecture. It consists of 30 WanAttentionBlock layers, each integrating vision self-attention, text-to-vision cross-attention, and feed-forward mechanisms. Multi-frame video inputs are encoded into a 3D vision latent by the 3D VAE encoder. These 3D vision latents are then flattened into a sequence and fed into the self-attention layers to encode temporal consistency. Finally, the sequence interacts with text tokens through cross-attention to achieve text-guided video generation. The self-attention mechanism captures intra-frame dependencies, while the cross-attention module integrates text tokens to enable text-guided video generation. This joint modeling of visual and textual information ensures high-quality, semantically aligned video outputs.

\textbf{Wan-Edit.} This elegant and streamlined architecture closely resembles that of image generation models, enabling FlowEdit to be directly integrated without significant modifications.
For multi-frame inputs from $i_1$ to $i_2$, denote as 
$\mathbf{X}_1^{src} = [X_1^{src, {i_1}}, \cdots, X_1^{src, {i_2}}]$, the editing process, as shown in Fig. \ref{fig:rf-edit} (b), handles all frames simultaneously by extending FlowEdit to the entire sequence, ensuring temporal consistency and efficient processing. The editing process follows the same formulation as Eq. (\ref{eq: flowedit}) and is defined as:
\begin{align}
    \label{eq: wanedit}
    \mathbf{X}_t^{edit} \!\!= \mathbf{X}_{t-1}^{edit} + \!\! \mathbb{E}[v^\Delta_t(\hat{\mathbf{X}}_t^{tgt}, \mathbf{X}_t^{src})|\mathbf{X}_1^{src}]dt,
\end{align}
where $\hat{\mathbf{X}}_t^{tgt} \approx \mathbf{X}_{t-1}^{edit}\!+\!\mathbf{X}_t^{src}\!-\! \mathbf{X}_1^{src}$.
This allows for smooth and coherent edits across all frames, preserving the temporal dynamics of the video.
\section{Experiments}
\label{sec:exp}

\subsection{Experimental  Settings}

\textbf{Baseline methods.} We compare our two RF-based editing methods with five diffusion-based models: TokenFlow \cite{geyer2024tokenflow}, DMT \cite{yatim2024space}, VidToMe \cite{li2024vidtome}, AnyV2V \cite{ku2024anyv2v}, and VideoGrain \cite{yang2025videograin}. Diffusion-based models require inversion before editing, while our RF-based methods are inversion-free. DMT relies on feature optimization of spatial marginal mean (SMM) during editing, and VideoGrain uses depth maps and object masks to assist editing.

\textbf{Evaluation metrics.} 
We conduct a fair evaluation of all models on our FiVE benchmark using 15 metrics. The ten commonly used metrics in Table \ref{tab:basellines} cover six aspects: structure distance~\citep{tumanyan2022splicing}, background preservation (PSNR, LPIPS~\citep{zhang2018unreasonable}, MSE, and SSIM~\citep{wang2004image} outside the editing mask), edit prompt-image consistency (CLIPSIM~\citep{clipsim} for the full image and masked regions), image quality assessment (NIQE~\cite{saad2012niqe}), temporal consistency (motion fidelity score~\cite{yatim2024space}) and running time. Additionally, our proposed FiVE-Acc introduces five metrics in Table. \ref{tab:basellines_five_acc} to evaluate the accuracy of successful object editing.

\textbf{Implementation details.} Pyramid-Edit utilizes the 384P Pyramid-Flow model, while Wan-Edit is based on the Wan2.1 1.3B model. Pyramid-Edit processes the first frame with 20 timesteps and subsequent frames with 10 timesteps per stage, across a total of three stages, whereas Wan-Edit operates with 50 timesteps throughout. For video editing, we follow the FlowEdit setting, where approximately the initial one-third of timesteps are skipped to achieve better results. Specifically, Pyramid-Edit skips all timesteps in the first stage, while Wan-Edit skips the first 15 timesteps. The source and target classifier-free guidance (CFG) are set to 7/5 (first/subsequent frames) and 10 in Pyramid-Edit, and 5 and 12 in Wan-Edit. Due to space constraints, the implementation details of the comparison methods are provided in the \textbf{Supplementary Materials}.
Considering the differences in the number of video frames each method can process, all DM-based methods except AnyV2V handle only the first 40 frames, while AnyV2V processes only the first 32 frames. RF-based methods process the first 41 frames due to the requirements of Video/3D VAE. All experiments were conducted on a single H100 GPU.

Please refer to the \textbf{Supplementary Materials} for details on baseline models, evaluation metrics, and implementation of the comparison methods.

\subsection{Quantitative Comparison}
Tables \ref{tab:basellines} and \ref{tab:basellines_five_acc} present the comparison results averaged across six editing types, covering both commonly used metrics and our proposed FiVE-Acc metrics. The results for each individual editing type are provided in the \textbf{Supplementary Materials}. To improve evaluation efficiency, all metrics except Image Quality Assessment (IQA) and Motion Fidelity Score are calculated by sampling one frame every eight frames. The final score is obtained by averaging the per-frame scores. This sampling method significantly reduces computational cost while maintaining a comprehensive evaluation to a certain extent. 

\subsubsection{Comparison on Common Metrics} 
In terms of background preservation, VideoGrain and Wan-Edit rank as the top two, exhibiting comparable performance across all metrics. Specifically, VideoGrain achieves the best results in Structure Distance (12.40) and two Background Preservation metrics (PSNR: 27.05, MSE: 25.10). Meanwhile, Wan-Edit, our proposed method, outperforms in the other two Background Preservation metrics (LPIPS: 94.61, SSIM: 82.55). For text-video similarity, VidToMe and DMT achieve the best results for both global (CLIPS.: 26.84) and edited-region text-video similarity (CLIPS.$_\text{edit}$: 21.44). Our proposed Pyramid-Edit and Wan-Edit follow closely, securing the second-best results with 26.82 and 21.23, respectively. 
For image quality assessment (IQA), the top three methods are TokenFlow, VideoGrain, and VidToMe, all of which are based on Stable Diffusion (SD) for image generation. This can be attributed to the inherent capability of SD models to generate high-quality images.

For temporal consistency, VidToMe and TokenFlow rank first and third in Motion Fidelity Score by propagating similarity between temporal tokens. Our proposed method, Wan-Edit, also maintains strong temporal consistency, securing second place. Finally, RF-based methods are significantly faster than diffusion-based models, with Pyramid-Edit and Wan-Edit achieving 1.44s and 3.07s per frame, compared to 25.98s and 27.12s for DMT and VideoGrain. This is mainly due to RF-based approaches leveraging Video/3D VAE, which enables higher temporal compression in the latent space, reducing the number of processed frames. For instance, Pyramid-Edit and Wan-Edit adopt temporal downsampling rates of 8× and 4×, respectively. Moreover, unlike diffusion-based methods, the inversion-free Pyramid-Edit and Wan-Edit bypass the inversion process, halving the runtime. Additionally, they require no extra conditions, such as depth maps or masks, further accelerating the editing process.

Overall, diffusion-based VideoGrain and our RF-based Wan-Edit perform similarly. However, VideoGrain relies on depth maps and object masks for editing and is significantly slower (27.12s per frame), whereas the inversion-free Wan-Edit achieves the highest efficiency with the lowest time (3.07s per frame). These findings clarify each method's strengths and trade-offs, guiding selection for specific applications and advancing video editing.

\begin{figure*}[t!]
    \centering
    \includegraphics[width=0.98\linewidth]{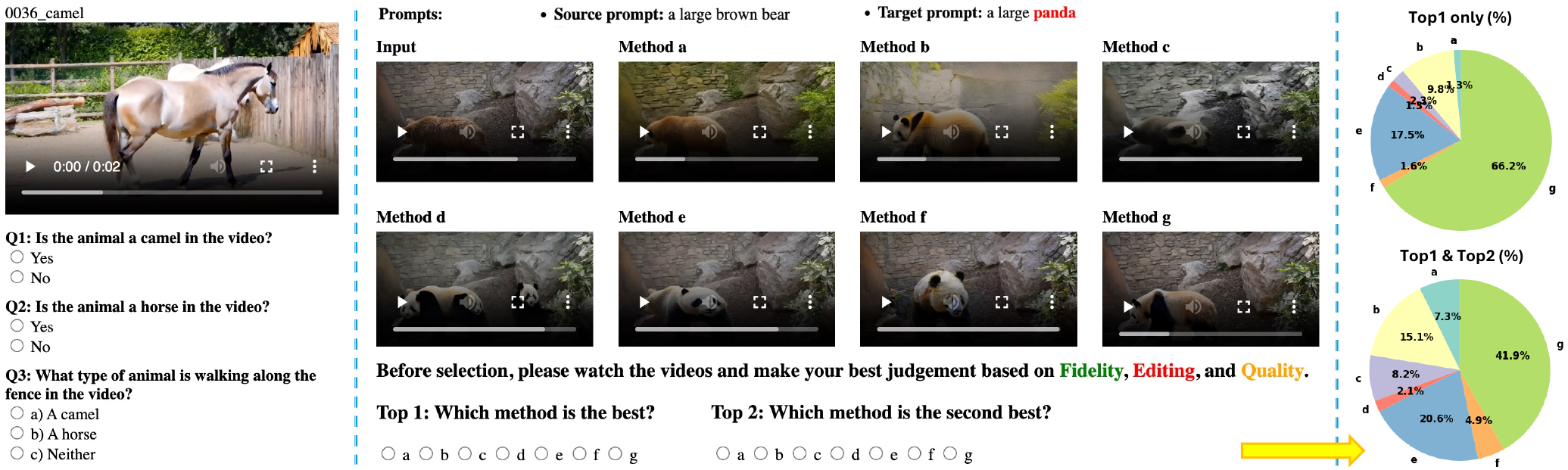}
    \vspace{-3mm}
    \captionof{figure}{Human evaluation example using \href{https://www.netlify.com/}{Netlify}. \textit{Left:} An example illustrating human verification of FiVE-Acc metric, conducted on WanEdit results. \textit{Central:} A human preference study where annotators select the top-2 preferred results. \textit{Right:} preference statistics.
    } \label{fig:human_eval}
\end{figure*}
\begin{figure}[ht]
    \centering
    \includegraphics[width=0.95\linewidth]{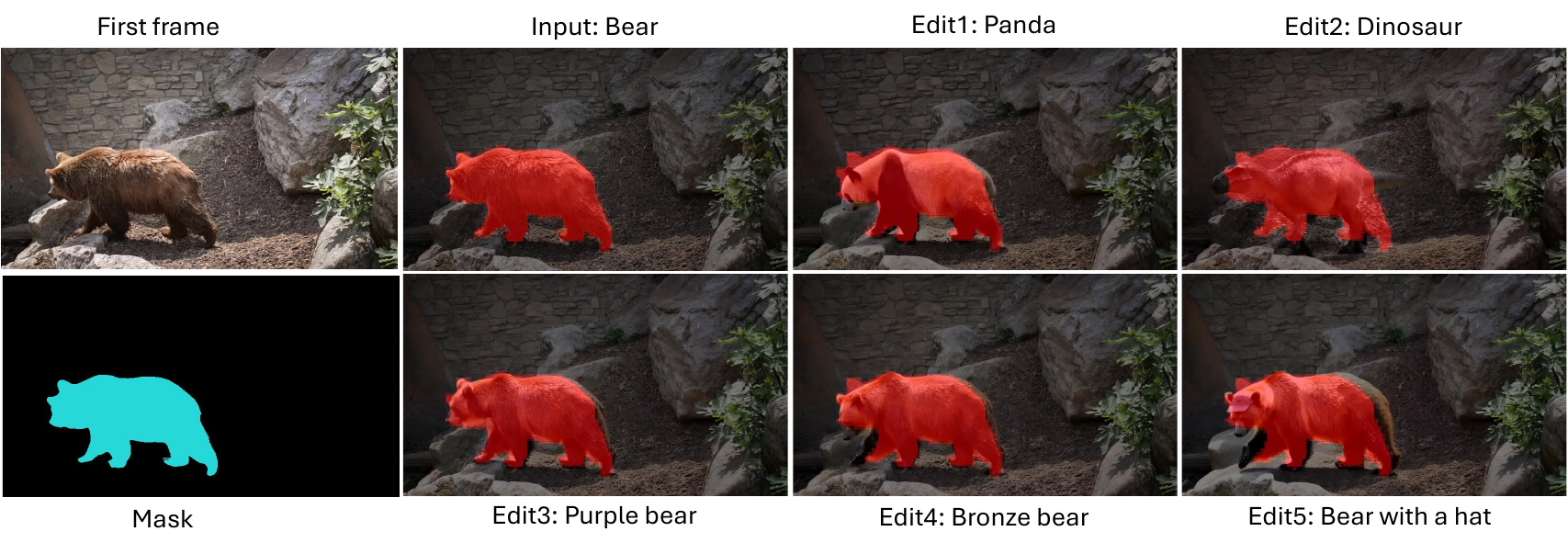}
    \vspace{-3mm}
    \caption{Human validation of mask quality.}
    \label{fig:masks}
\end{figure}

\begin{figure*}
    \centering
    \includegraphics[width=0.98\linewidth]{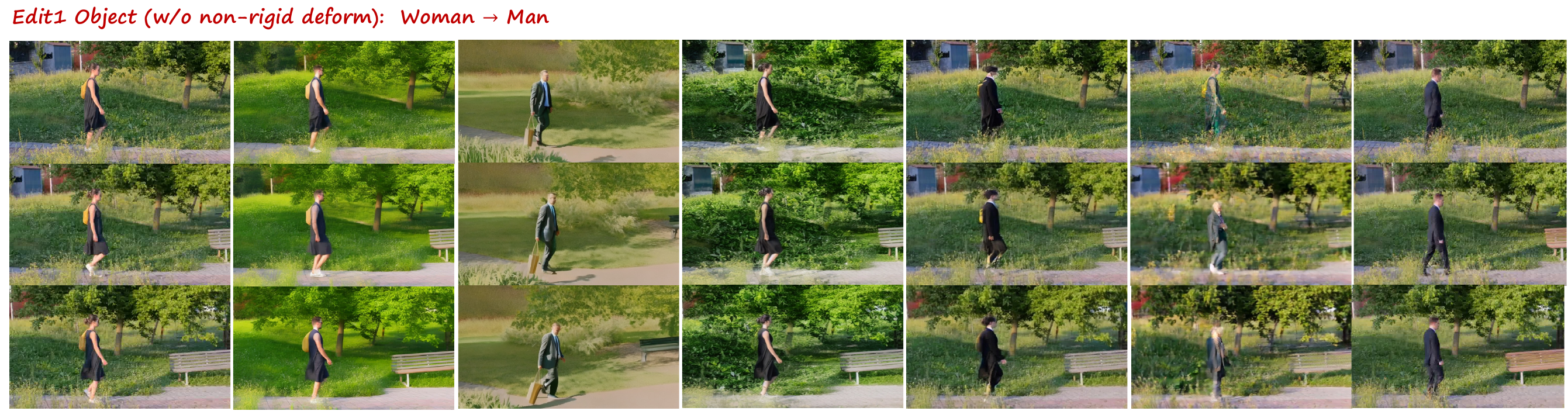}
    \includegraphics[width=0.98\linewidth]{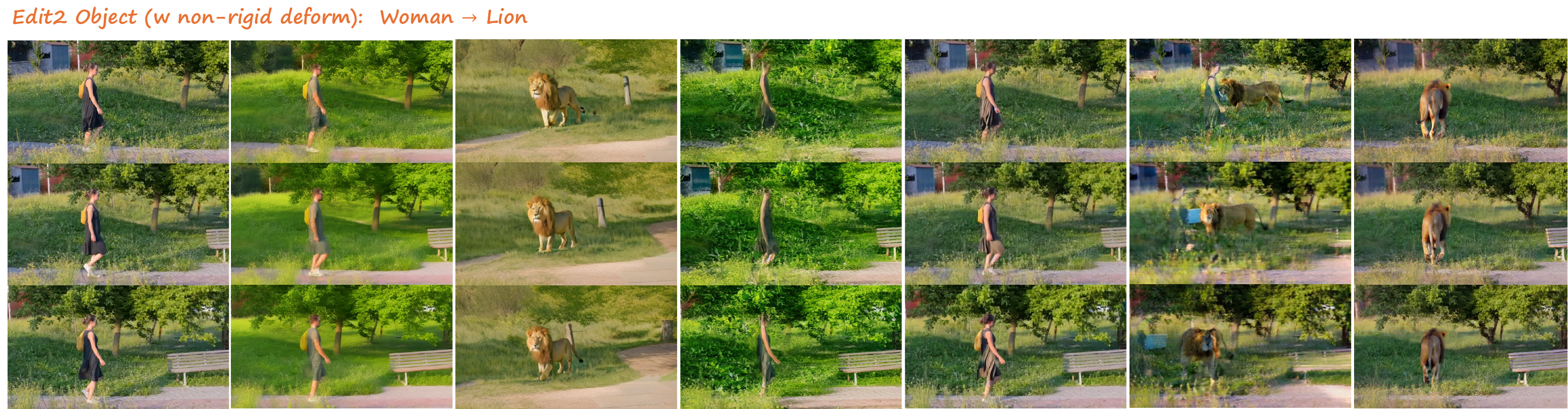}
    \includegraphics[width=0.98\linewidth]{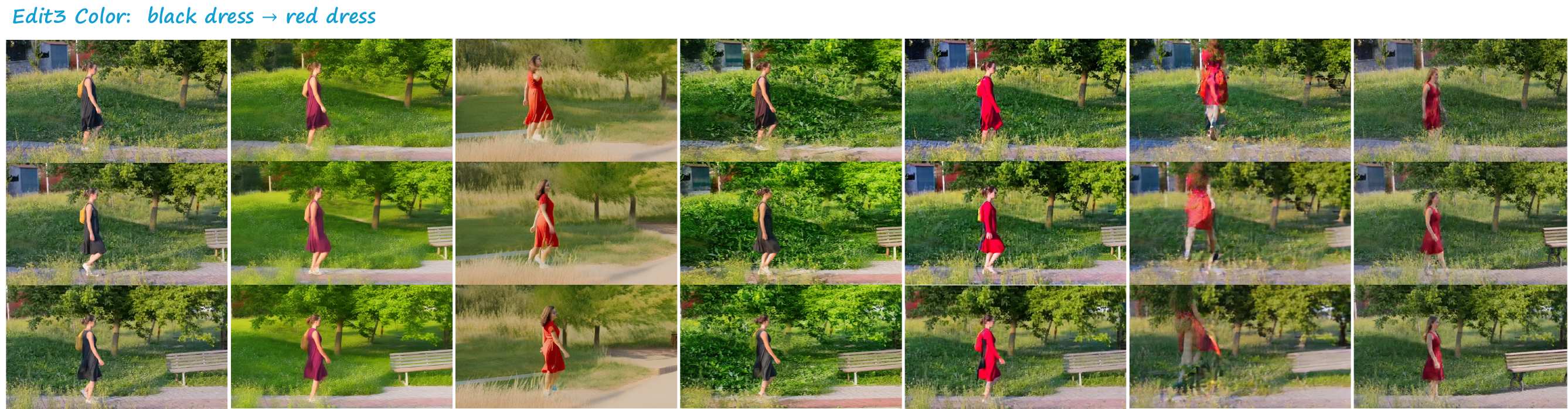}
    \includegraphics[width=0.98\linewidth]{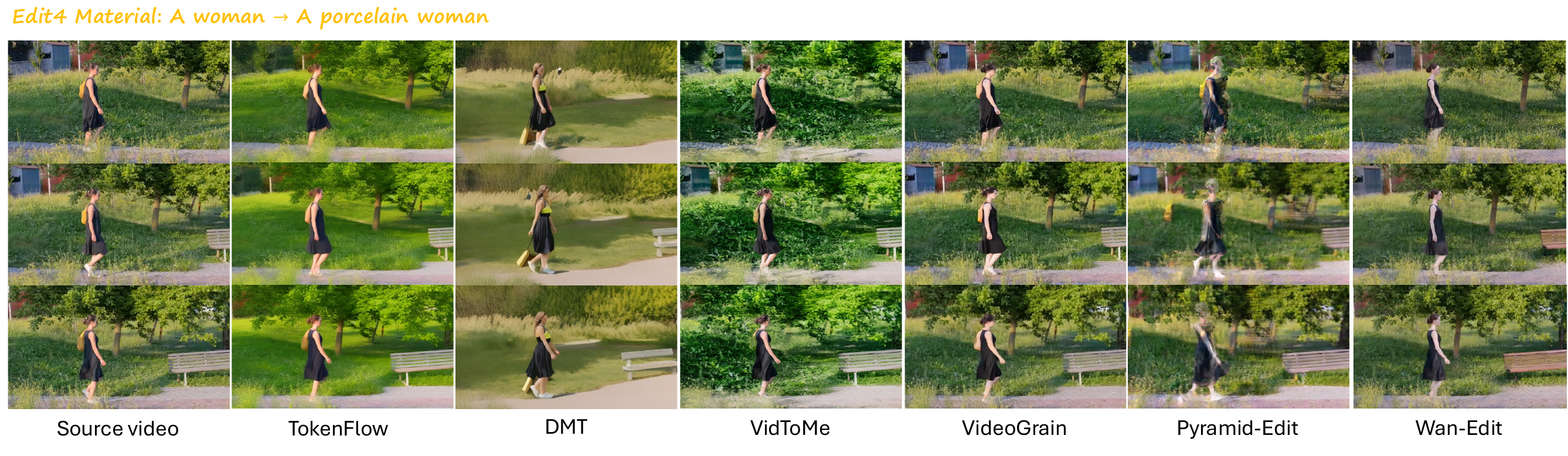}
    \includegraphics[width=0.975\linewidth]{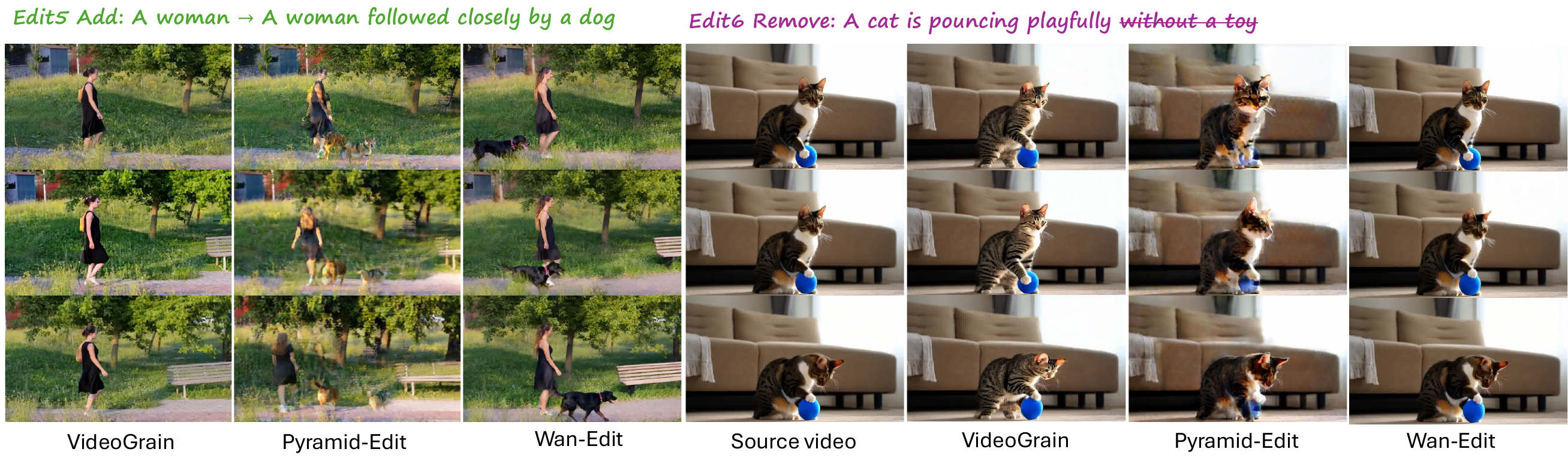}
    \vfill
    \vspace{-3mm}
    \caption{Editing results across six editing types and five comparison methods.}
    \label{fig:vis_results}
\end{figure*}

\subsubsection{Comparison on FiVE-Acc Metrics} 
Notably, the text-video similarity scores in Table \ref{tab:basellines} for all methods show only minor variations, with global CLIP scores (CLIPS.) clustering around 26 and edited-region CLIP scores (CLIPS.$_\text{edit}$) around 21. This narrow gap suggests that CLIP score may lack the sensitivity needed to comprehensively evaluate fine-grained text-video alignment in editing tasks. To overcome this limitation, our proposed VLM-based FiVE-Acc metrics, presented in Table \ref{tab:basellines_five_acc}, leverage vision-language models to better capture semantic changes in target objects, revealing key insights and enabling a more fine-grained evaluation.

Table \ref{tab:basellines_five_acc} presents the accuracy results averaged across six editing types. Overall, the FiVE-Acc metrics yield conclusions similar to the CLIP score, indicating that DMT, Wan-Edit, and Pyramid-Edit outperform other methods. However, the variance among FiVE-Acc metrics is notably higher. For instance, while TokenFlow and VidToMe achieve scores close to DMT in CLIP score evaluation, their FiVE-Acc accuracy is only about half of DMT’s—27.43\% and 26.77\% compared to 48.42\%. This discrepancy arises because both TokenFlow and VidToMe rely on feature clustering to propagate similar tokens across frames, making them ineffective for non-rigid transformations. For example, in Fig. \ref{fig:vis_results}, when editing a woman into a lion, these methods retain the rigid structure of human being while altering only attributes like texture and color to resemble a lion.

A detailed comparison and analysis of the FiVE-Acc metrics is provided. For Yes/No questions (FiVE-YN), our Wan-Edit and the diffusion-based DMT with optimization rank the highest, achieving 41.41\% and 34.78\%, respectively. In multi-choice questions (FiVE-MC), DMT and our Pyramid-Edit perform best, with accuracy rates of 62.06\% and 54.01\%, respectively. FiVE-$\cup$ represents cases where at least one of FiVE-YN or FiVE-MC holds true. The highest accuracy is again achieved by DMT (62.98\%) and our Pyramid-Edit (56.36\%), indicating consistency between FiVE-YN and FiVE-MC evaluations. Conversely, FiVE-$\cap$ requires both FiVE-YN and FiVE-MC to hold simultaneously. Here, our Wan-Edit and DMT obtain the highest accuracy, reaching 38.22\% and 33.86\%, respectively. The strong performance of Wan-Edit in FiVE-$\cap$ suggests that it excels in specific target edits while maintaining minimal deviation from real-world object distributions. Finally, for the overall FiVE-Acc metric, which averages the four aforementioned metrics, DMT and Wan-Edit achieve the top two results, with accuracy rates of 48.42\% and 46.97\%, respectively. However, DMT is an optimization-based method that requires per-prompt optimization for each video, whereas Wan-Edit operates without any additional requirements. 

Overall, our proposed VLM-based FiVE-Acc evaluation is essential for assessing semantic fidelity, especially in fine-grained editing. It can effectively capture subtle semantic changes in target objects that traditional metrics like CLIP score may overlook.

\subsubsection{Comparison on All Metrics}
Overall, as shown in Tables \ref{tab:basellines} and \ref{tab:basellines_five_acc}, RF-based video editing methods achieve comparable or superior background preservation, text-video similarity, editing accuracy, and motion consistency compared to diffusion-based methods, while being 10–15 times faster than diffusion-based models, such as VideoGrain. In terms of IQA metrics, since fine-grained video editing focuses on modifying only the target object while preserving the background, our RF-based methods, Pyramid-Edit and Wan-Edit, maintain IQA scores nearly identical to those of the source video, rather than artificially enhancing the edited frames' quality as seen in other methods. This suggests that our approach ensures minimal distortion to the original video while achieving precise and semantically aligned edits. 

Moreover, our results indicate that diffusion-based methods often require additional guidance, such as depth maps or masks, to ensure accurate editing, whereas our RF-based methods, particularly Pyramid-Edit and Wan-Edit, operate efficiently without such dependencies. The efficiency of RF-based approaches is further attributed to the inversion-free pipeline and Video/3D VAE, which enable higher temporal compression in the latent space, significantly reducing computational costs while maintaining high-quality edits.

\subsection{Human Validation}
\begin{table}[t!]
    \centering
    \caption{Human vs. VLM model evaluation on FiVE-Acc metrics.}
    \label{tab:human_eval_five_acc}
    \vspace{-2mm}
    {\footnotesize
    \renewcommand{\arraystretch}{0.8} 
    \setlength{\tabcolsep}{2.35pt}
    \begin{tabular}{c|ccccc}
        \toprule
        Evaluator & {FiVE-YN} &{FiVE-MC} &{FiVE-$\cup$} &{FiVE-$\cap$} & FiVE-Acc$\uparrow$ \\
        \midrule
        Qwen-2.5-VL &41.41 &52.53 &55.72 &38.22 &46.97 \\
         Human      &44.37 &50.31 &51.98 &42.70 &47.34 \\
        \bottomrule
    \end{tabular}
    }
\end{table}

To ensure the reliability and perceptual alignment of our evaluation protocol, we conduct comprehensive human validation on the proposed FiVE-Acc metric, the compared editing methods, and the quality of segmentation masks.

\textbf{Human validation of FiVE-Acc metrics.}
In Fig. \ref{fig:human_eval} (left), we conducted a human study on 16 randomly sampled videos with 64 prompts. Table~\ref{tab:human_eval_five_acc} shows that human scores closely match Qwen-2.5-VL's results (47.34 vs. 46.97), confirming FiVE-Acc aligns well with human perception and offers consistent evaluation.

\textbf{Human validation across compared methods.}
In Fig. \ref{fig:human_eval} (central), we randomly sample 11 videos, resulting in 45 source–target prompt pairs. All seven methods from main Table~\ref{tab:basellines} (methods a–g) are evaluated. To reduce bias, the order of methods is randomized. Annotators are asked to select their top-2 preferred results per prompt, and the aggregated votes are used to compute preference statistics in Fig. \ref{fig:human_eval} (right). The results indicate that Method \textbf{g} (our Wan-Edit) significantly outperforms the others, receiving 66.2\% top-1 votes and 41.9\% top-1\&2 combined votes.

\textbf{Human validation of mask quality.}
The segmentation masks in FiVE are initially generated by the SAM model to provide object-level supervision. To ensure accuracy and alignment with the intended target regions, each mask is manually reviewed and corrected if necessary by human annotators (see Fig.~\ref{fig:masks}). This process guarantees high-quality annotations that support reliable evaluation and training.

\subsection{Qualitative Comparison}
Fig. \ref{fig:vis_results} compares the editing results of all methods across six editing types. More visual comparisons can be found in the appendix or on our project page.\footnote{\url{https://sites.google.com/view/five-benchmark}} Analyzing by editing type, object edits without non-rigid transformations ({\color{BrickRed} Edit1}) and color changes ({\color{cyan} Edit3}) are the easiest, successfully handled by nearly all methods. In contrast, object edits with non-rigid transformations ({\color{orange} Edit2}) and material changes ({\color{Yellow} Edit4}) are only effectively performed by the optimization-based DMT and our Wan-Edit. Object addition ({\color{Green} Edit5}) is successfully achieved only by Wan-Edit, while object removal ({\color{Fuchsia} Edit6}) fails across all methods—only Pyramid-Edit manages partial removal, but with suboptimal results.

From the perspective of background preservation in fine-grained video editing, DMT and VidToME introduce significant alterations, while TokenFlow, VideoGrain, and the RF-based Pyramid-Edit and Wan-Edit largely retain the original background information. Notably, the two diffusion-based methods enhance background quality, benefiting from Stable Diffusion's high-quality image generation model. For foreground object editing, TokenFlow, VidToME, and VideoGrain fail to handle non-rigid structural changes. Pyramid-Edit, due to its multi-resolution processing along timesteps and autoregressive temporal architecture, suffers from noise intensity discrepancies and accumulated temporal errors. This results in overlapping or blurring between the source and target objects, making it less suitable for video editing tasks. Both DMT and Wan-Edit achieve strong target object editing, but Wan-Edit preserves object detail and pose consistency more effectively.

The qualitative analysis further supports the quantitative results: DMT excels in object editing but distorts the background, whereas VideoGrain and Wan-Edit achieve the best balance between foreground and background preservation. More quantitative and qualitative comparison on each fine-grained editing type can be found in the \textbf{Supplementary Materials}.

\subsection{Limitations and Future Work.} 
This work adapts image editing techniques~\ref{eq: flowedit} to video editing, demonstrating strong performance on the Wan2.1 T2V model, which shares the same architecture as I2V methods. However, it is less suited for the pyramid structure of Pyramid-Flow T2V model, leaving room for improvement in editing quality. In future work, we aim to further refine RF-based video editing models to ensure broader applicability across different architectures. Additionally, addressing challenging cases in benchmark tests—such as large motions, object removal, and long videos—will be a key focus for improvement.
\section{Conclusion}
We introduce FiVE, a benchmark for fine-grained video editing, and propose the VLM-based FiVE-Acc metric, which evaluates the accuracy of object-level editing success. Additionally, we adapt two RF-based video editing methods, Pyramid-Edit and Wan-Edit. To the best of our knowledge, this is the first comprehensive quantitative and qualitative comparison of emerging diffusion-based and flow-based models in the video editing community. FiVE and FiVE-Acc provide a standardized framework for evaluating fine-grained video editing models, guiding future advancements in efficient and high-fidelity video editing solutions. We hope this benchmark drives innovation in both research and real-world applications.

{
    \small
    \bibliographystyle{ieeenat_fullname}
    \bibliography{main}
}

\newpage
\newpage
\setcounter{section}{0} 
\twocolumn[ 
  \centering
  \vspace{2em} 
  \textbf{\LARGE Supplementary Materials} \\ 
  \vspace{3em} 
]
\renewcommand\thesection{\Alph{section}}

In this supplementary file, we provide the following materials:
\begin{itemize}
    \item More details on the baseline methods
    \item More details on implementation details 
    \item More details on FiVE Dataset 
    \item GPU memory and speed comparison  
    \item More quantitative results and analysis
    \item More qualitative results and analysis
\end{itemize}

\section{Baseline Methods}
\begin{table*}[h]
    \centering
    \caption{Comparison of different video editing methods for DM and RF models under default settings.}
    \label{tab:methods}
    \resizebox{0.99\textwidth}{!}{
    \renewcommand{\arraystretch}{1.3} 
    \setlength{\tabcolsep}{3pt} 
    \begin{tabular}{cl|ccccccc|ccc}
        \hline
        \multicolumn{2}{c|}{\multirow{2}{*}{\textbf{Methods}}} & \multirow{2}{*}{\textbf{Publication}} & \textbf{Inv.} & \textbf{Attn} & \textbf{Base T2I/V} & \multirow{2}{*}{\textbf{Inv.-free}} & \multirow{2}{*}{\textbf{Resolution}}  & \textbf{Timesteps} & \multirow{2}{*}{\textbf{Conditions}} \\ 
        & &&\textbf{Type} &\textbf{Injection} & \textbf{Model} & &&\textbf{Inv.+Edit}\\
        \hline
        \multirow{5}{*}{DMs} 
        &{TokenFlow}  & ICCV23 &DDIM~\cite{song2021denoising}  &\cmark  & SD2.1~\cite{rombach2022high} & \xmark & (512, 512) & 500 + 50 & \xmark \\ 
        & {DMT} & CVPR24 & DDIM~\cite{song2021denoising} &\xmark  & ZeroScope~\cite{zeroscope} & \xmark & (576, 320) & 1000 + 50  & Optimization\\ 
        & {Vidtome}  & CVPR24 & PnP~\cite{tumanyan2023plug} & \cmark & SD1.5~\cite{rombach2022high} & \xmark & (512, 512)  & 50 + 50 & \xmark  \\ 
        & {AnyV2V} & TMLR24 & PnP~\cite{tumanyan2023plug} &\cmark  &I2VGen-XL~\cite{zhang2023i2vgen} & \xmark & (512, 512)  &500 + 50 & InstrctPix2Pix~\cite{brooks2023instructpix2pix} \\ 
        &VideoGrain &ICLR25 &DDIM~\cite{song2021denoising} &\cmark &SD1.5~\cite{rombach2022high} & \xmark &(512, 512) &50 + 50 & Depth + Mask\\
        \hline
        \multirow{2}{*}{RFs} 
        & Pyramid-Edit & Ours & FlowEdit~\cite{kulikov2024flowedit} &\xmark  & Pyramid-Flow~\cite{jin2024pyramidal} & \cmark & (640, 384)  & 40 & \xmark \\ 
        & Wan-Edit & Ours & FlowEdit~\cite{kulikov2024flowedit} &\xmark  & Wan2.1~\cite{wan2025wan} & \cmark &(832, 480) & 50 & \xmark \\ 
        \hline
    \end{tabular}
    }
\end{table*}

\begin{figure*}[t]
    \centering
    \includegraphics[width=0.95\textwidth]{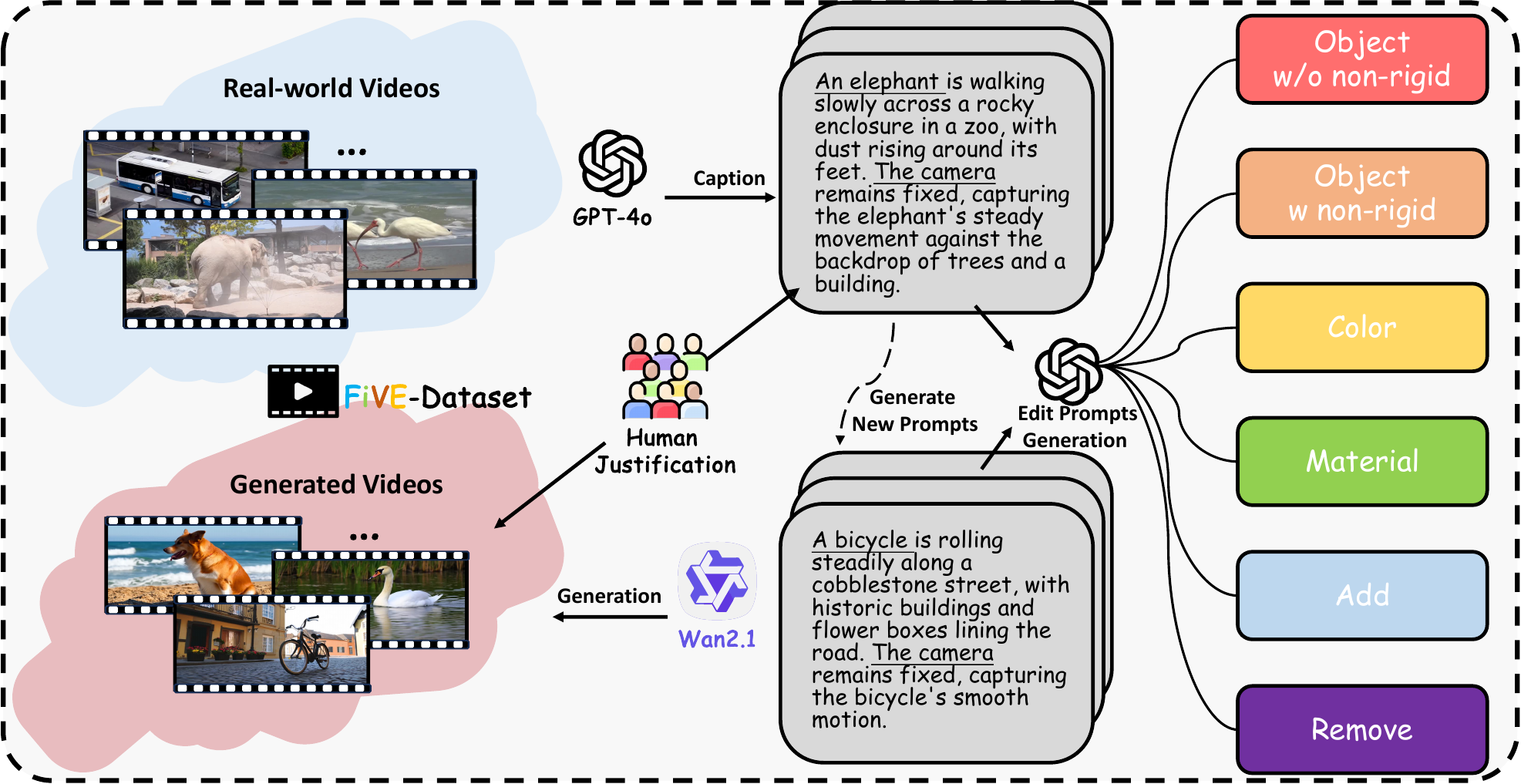}
    \vspace{-2mm}
    \caption{FiVE-Dataset construction pipeline.}
    \label{fig:datapipe}
\end{figure*}
\begin{figure*}[h]
    \centering
    \includegraphics[width=0.99\textwidth]{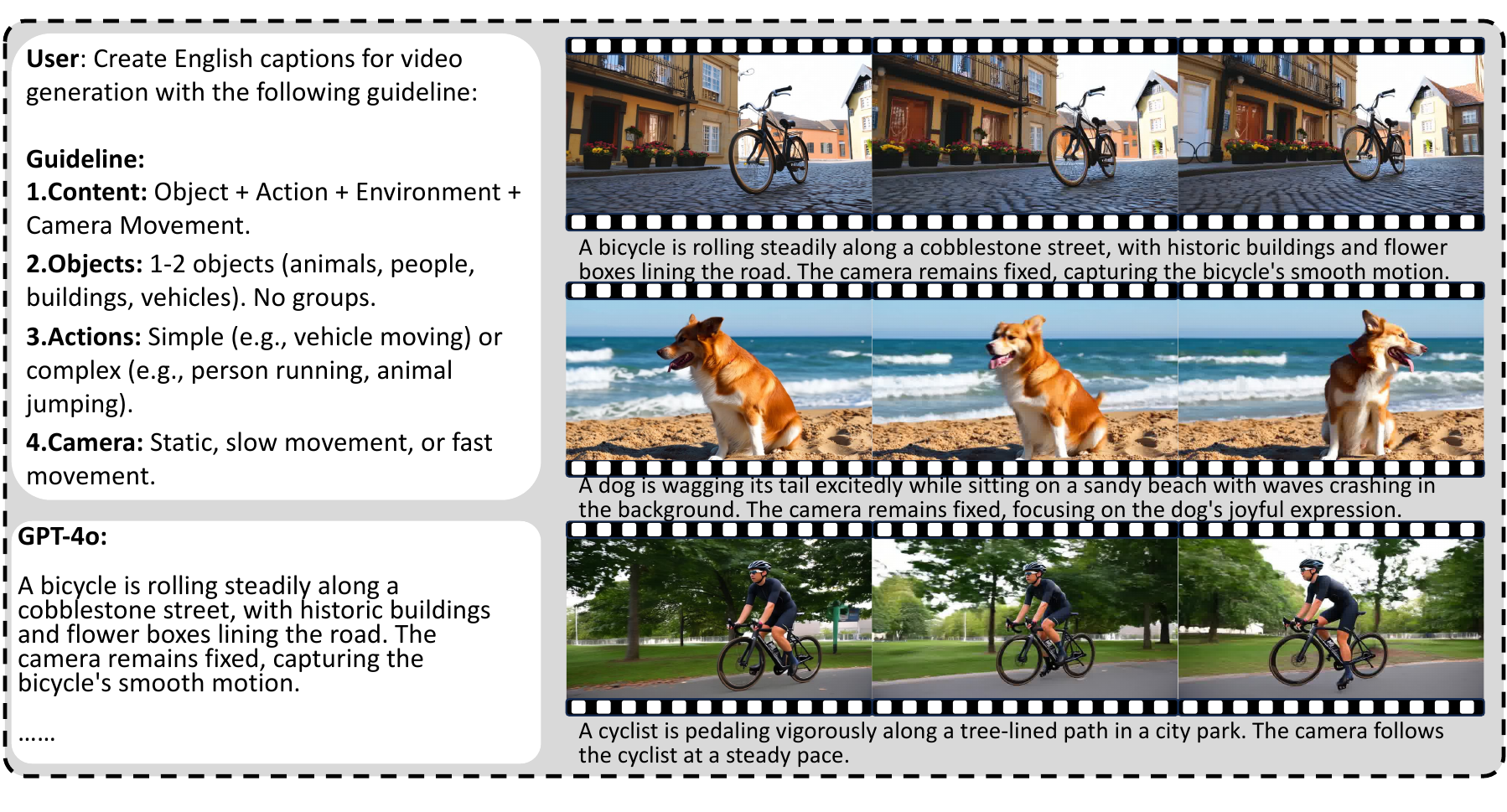}
    \vspace{-4mm}
    \caption{Example of generated caption-video pairs.}
    \label{fig:gen_video}
\end{figure*}

\begin{itemize}
    \item TokenFlow~\cite{geyer2024tokenflow} is a training-free framework for consistent video editing that leverages diffusion features by enforcing cross-frame semantic token alignment in latent space to preserve spatiotemporal coherence. By propagating consistent appearance and motion patterns through optimized token interactions in a pre-trained text-to-image model, it achieves temporally stable edits without requiring additional fine-tuning or annotated data.
    \item DMT~\cite{yatim2024space} is a zero-shot framework for text-driven diffusion motion transfer that leverages spatiotemporal diffusion features to align source motion patterns with target textual descriptions in a unified latent space. By integrating cross-modal attention mechanisms and temporal coherence constraints within a pre-trained diffusion model, it enables realistic motion synthesis without requiring task-specific training or paired data, ensuring both semantic fidelity and dynamic consistency.
    \item VidToME~\cite{li2024vidtome} is a zero-shot video editing framework that enhances spatiotemporal consistency by adaptively merging redundant tokens across frames within a pre-trained diffusion model. This token-efficient strategy preserves critical motion and appearance features while reducing computational overhead, enabling coherent video edits without task-specific training or temporal-aware fine-tuning.
    \item AnyV2V~\cite{ku2024anyv2v} is a tuning-free framework designed for universal video editing tasks, leveraging spatiotemporally consistent diffusion features through cross-frame latent propagation to maintain coherence across diverse editing operations. By dynamically aligning semantic and motion patterns in pre-trained diffusion models without task-specific tuning, it enables flexible video-to-video transformations while preserving temporal stability and visual fidelity. For prompt-based editing, it uses InstructPix2Pix~\cite{brooks2023instructpix2pix} to edit the first frame first.
    \item VideoGrain~\cite{yang2025videograin} is a video editing framework that enables multi-grained control through hierarchical space-time attention modulation, dynamically adjusting spatial and temporal feature interactions in diffusion models to achieve precise edits across varying granularities. By decomposing and recombining cross-frame attention patterns at different resolution scales, it maintains temporal coherence and visual fidelity while supporting diverse editing tasks without requiring architectural modifications or task-specific fine-tuning.
\end{itemize}

\section{Implementation Details}
All experiments were conducted using the official GitHub repository and environment, with default settings. For training-free methods, the editing results are highly dependent on hyperparameters, such as in PNP~\cite{tumanyan2023plug}. To minimize the impact of hyperparameters, we randomly selected six videos from our benchmark and performed a search within an appropriate parameter range to find the best hyperparameters. These were then fixed for all subsequent data in the benchmark. All experiments were run on a single H100 GPU. Table \ref{tab:methods} lists the parameter settings for the compared methods and our proposed approach. 

For VLM-based FiVE-Acc evaluation, QWen2.5-VL-7B is selected as the evaluation model. We sample one frame every 8 frames from the edited video, selecting a total of 5 evenly spaced frames from a 40-frame video, which are then fed into the vision encoder of QWen2.5-VL. The text input consists of Yes/No questions or multiple-choice questions, as illustrated in Fig. 2 of the main paper. Considering the varying number of videos across different editing types, we compute the FiVE-Acc metric separately for each type. The final FiVE benchmark score is obtained by averaging the scores across all six editing types, as presented in Table 3.

\begin{figure*}
    \centering
    \includegraphics[width=0.99\textwidth]{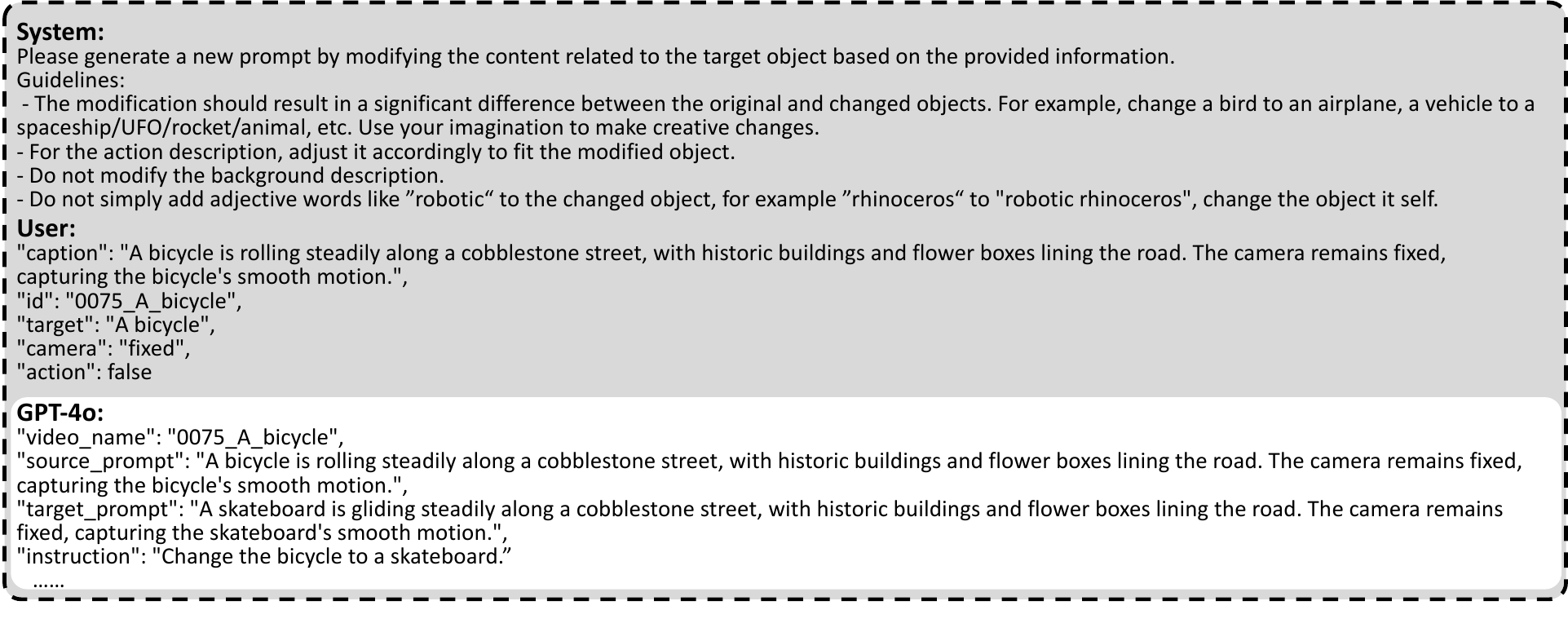}
    \vspace{-4mm}
    \caption{Example of editing prompt generation.}
    \label{fig:prompt_gen}
\end{figure*}

\begin{figure*}
    \centering
    \includegraphics[width=0.99\textwidth]{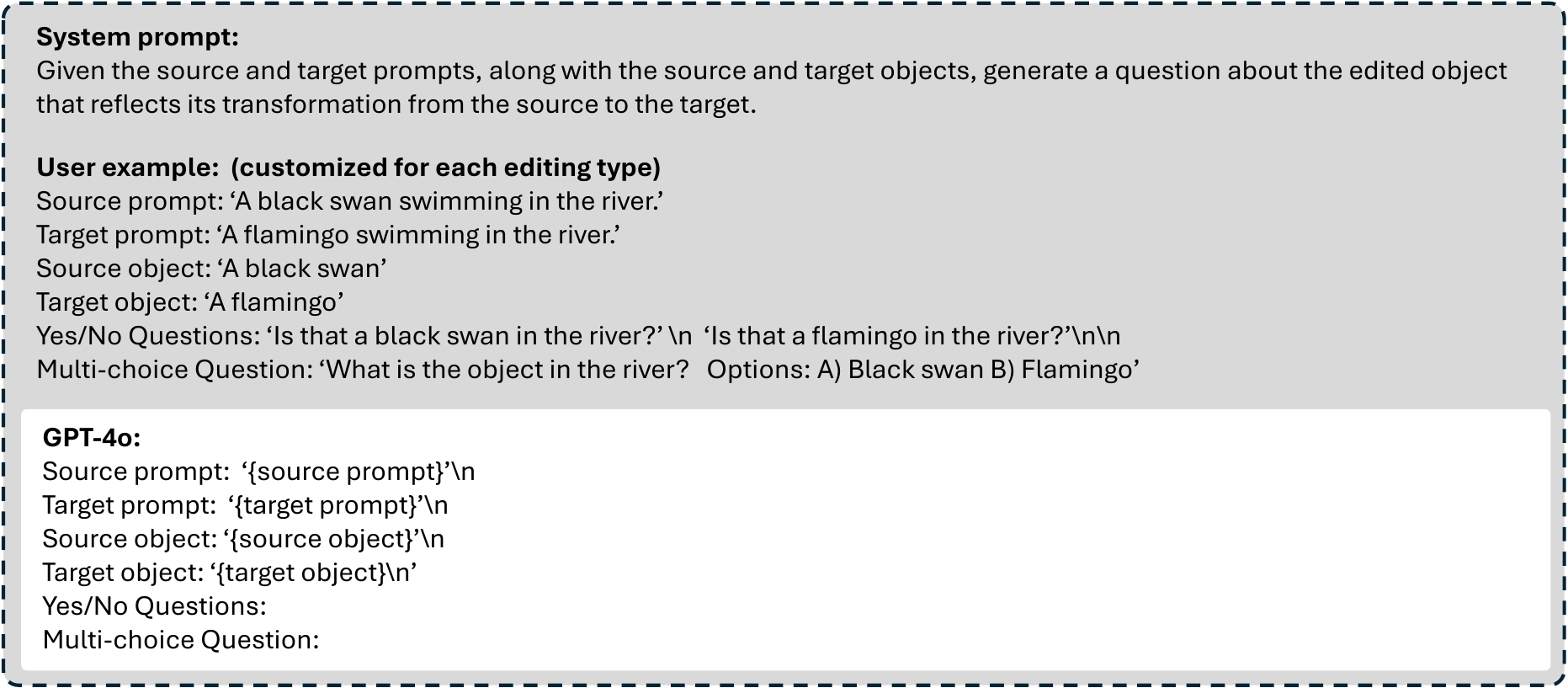}
    \vspace{-1mm}
    \caption{Example of Yes/No and Multi-choice question generation for FiVE-Acc evaluation.}
    \label{fig:question_gen}
\end{figure*}

\section{More Details on FiVE Dataset}
The construction of the FiVE-Dataset involves the collection of real-world videos, the generation of captions for these videos, the creation of synthetic video-caption pairs, and the generation of editing prompts. The overall pipeline is illustrated in~\cref{fig:datapipe}. 

\subsection{Video-description Pair Construction.} 

We begin by selecting real-world videos from the DAVIS dataset~\cite{perazzi2016davis} that are well-suited for fine-grained video editing. For each chosen video, we use GPT-4o~\cite{hurst2024gpt} to generate detailed annotations every 8th frame, capturing key elements such as subject actions, background details, and camera movements. Next, we create new annotations in the style of real video descriptions, which are then used to guide a text-to-video model in generating new videos. The full process and examples of generated pairs are shown in~\cref{fig:gen_video}. In this process, human justification is involved in assessing the quality of both the videos and their descriptions to ensure the generation of high-quality video-description pairs.

\subsection{Editing Prompt Generation}
For the constructed video-caption pairs, we design specialized prompts to generate target editing instructions for six editing types. GPT-4o is employed to create new video captions by modifying the original captions based on the target object, serving as the target prompts for the editing process. \cref{fig:prompt_gen} provides an example of the prompt and its corresponding output for generating an editing type instruction.

\subsection{FiVE-Acc Question Generation}  
The {FiVE benchmark} provides both the \textit{source object} (the object in the original video) and the \textit{target object} (the object after editing). Based on this information, we utilize {GPT-4o} to generate {Yes/No questions} and {Multiple-choice questions} to assess the accuracy of the edits, as shown in \cref{fig:question_gen}. The user prompt is customized for each editing type, and the generated questions are manually reviewed for quality assurance. These generated questions enable an automated evaluation of editing success based on the \textbf{FiVE-Acc} metric.

\section{GPU Memory and Speed}
\begin{figure}[t]
    \centering
    \includegraphics[width=0.99\linewidth]{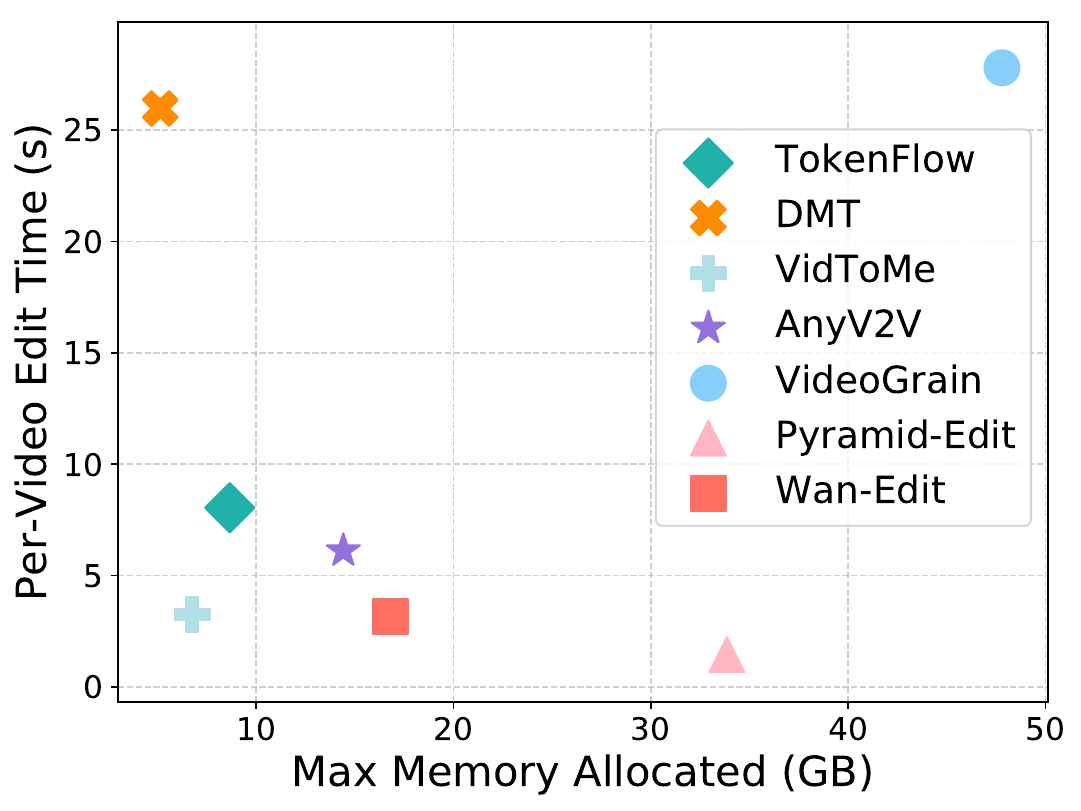}
    \caption{Comparison of editing efficiency, including GPU memory usage and per-frame running time. All test on a single NVIDIA H100.}
    \label{fig:gpu_time_comparison}
\end{figure}

As shown in Fig.~\ref{fig:gpu_time_comparison}, TokenFlow, VidToMe, AnyV2V, and our proposed Wan-Edit are all positioned in the lower-left corner, indicating a well-balanced trade-off between editing time and peak memory usage. However, Wan-Edit significantly outperforms the other three in terms of editing quality, further demonstrating its effectiveness. Compared to the highly competitive VideoGrain, Pyramid-Edit and Wan-Edit drastically reduces editing time and memory consumption, proving their efficiency.

The speed differences among these methods stem from their architectural choices and computational requirements. Pyramid-Edit is the fastest, benefiting from its multi-resolution design and the high spatiotemporal compression rate of VideoVAE (8×8×8), which significantly reduces the processing burden. However, this aggressive compression can lead to background collapse, particularly when the background exhibits fast motion. Wan-Edit, on the other hand, adopts a more moderate compression rate (4×8×8), which balances efficiency and quality, ensuring better background preservation while maintaining competitive speed. In contrast, VideoGrain is significantly slower due to its dependence on segmentation and depth models. These additional processing steps, which involve extracting object masks and depth maps to guide the editing process, introduce substantial computational overhead. This makes VideoGrain less suitable for real-time or high-speed applications despite its strong editing accuracy.

\section{More Quantitative Results and Analysis}
\begin{table*}[t]
    \centering
    \caption{{\color{BrickRed} Edit1}: Comparison of diffusion- and flow-based video editing methods for {\color{BrickRed} object replacement without non-rigid transformations} on the FiVE benchmark.}
    \resizebox{0.99\textwidth}{!}{
    \renewcommand{\arraystretch}{1.2}
    \setlength{\tabcolsep}{2.8pt} 
    \begin{tabular}{ll|c|cccc|cc|c|cc}
        \Xhline{1pt}
        \multicolumn{2}{c|}{\multirow{2}{*}{\textbf{Methods}}} & \textbf{{Structure}} &\multicolumn{4}{c|}{\textbf{Background Preservation}} &\multicolumn{2}{c|}{\textbf{Text Alignment}} &\multicolumn{1}{c|}{\textbf{IQA}} &\multicolumn{1}{c}{\textbf{Temp. Consis.}} \\
        & &Dist.$_{\times 10^3}$$\downarrow$ &PSNR$\uparrow$ &LPIPS$_{\times 10^3}$$\downarrow$ &MSE$_{\times 10^4}$$\downarrow$ &SSIM$_{\times 10^2}$$\uparrow$ &CLIPS.$\uparrow$ &CLIPS.$_{\text{edit}}\uparrow$ &NIQE$\downarrow$ & Motion Fidelity S.$_{\times 10^2}$$\uparrow$  \\
        \Xhline{1pt} 
        &Source Videos &0.00 & $\infty$ &0.00 &0.00 &100.00 &23.97 & 18.75 &6.33 &93.76 \\
        \Xhline{1pt} 
        \multirow{5}{*}{DMs} 
        &TokenFlow~\cite{geyer2024tokenflow} &36.24 & 19.12 & 257.32 & 137.12 & 72.63 & 27.04 & 21.23 &\textbf{4.05} &88.23 \\ 
        &DMT~\cite{yatim2024space}           &85.35 & 14.35 & 410.83 & 402.43 & 51.61 & 27.26 & \textbf{21.52} &5.25 &81.93 \\
        &VidToMe~\cite{li2024vidtome}        &23.81 & 21.32 & 261.19 & 88.47 & 71.50 & \textbf{27.75} & 21.19 &4.68  &\textbf{90.65} \\
        &AnyV2V~\cite{ku2024anyv2v}          &71.19 & 15.85 & 345.77 & 350.37 & 50.91 & 25.41 & 19.96 &4.64  &61.63 \\
        &VideoGrain~\cite{yang2025videograin} &\textbf{11.71} & \textbf{26.92} & \underline{184.08} & \textbf{26.82} & \underline{79.06} & \underline{27.43} & \underline{21.40} &\underline{4.10} &88.29 \\
        \hline
        \multirow{2}{*}{\makecell{RFs \\ (Ours)}} 
        &Pyramid-Edit  &28.27 & 20.87 & 276.18 & 96.15 & 72.56 & \underline{27.43} & 20.11 &5.47 &81.52 \\
        &Wan-Edit  &\underline{13.50} & \underline{24.81} & \textbf{93.67} & \underline{39.67} & \textbf{82.54} & 27.19 & 21.38 &6.59 &\underline{89.37} \\
        \Xhline{1pt}
    \end{tabular}}

    \label{tab:baselines-edit1}
\end{table*}

\begin{table*}[h]
    \centering
    \caption{{\color{orange} Edit2}: Comparison of diffusion- and flow-based video editing methods for {\color{orange} object replacement with non-rigid transformations} on the FiVE benchmark.}
    \label{tab:baselines-edit2}
    \resizebox{0.99\textwidth}{!}{
    \renewcommand{\arraystretch}{1.2}
    \setlength{\tabcolsep}{2.8pt} 
    \begin{tabular}{cl|c|cccc|cc|c|cc}
        \Xhline{1pt}
        \multicolumn{2}{c|}{\multirow{2}{*}{\textbf{Methods}}}  & \textbf{{Structure}} &\multicolumn{4}{c|}{\textbf{Background Preservation}} &\multicolumn{2}{c|}{\textbf{Text Alignment}} &\multicolumn{1}{c|}{\textbf{IQA}} &\multicolumn{1}{c}{\textbf{Temp. Consis.}} \\
        & &Dist.$_{\times 10^3}$$\downarrow$ &PSNR$\uparrow$ &LPIPS$_{\times 10^3}$$\downarrow$ &MSE$_{\times 10^4}$$\downarrow$ &SSIM$_{\times 10^2}$$\uparrow$ &CLIPS.$\uparrow$ &CLIPS.$_{\text{edit}}\uparrow$ &NIQE$\downarrow$ &Motion Fidelity S.$_{\times 10^2}$$\uparrow$  \\
        \Xhline{1pt} 
        &Source Videos &0.00 & $\infty$ &0.00 &0.00 &100.00 &22.13 & 17.33 &6.33 &93.76 \\
        \Xhline{1pt}
        \multirow{5}{*}{DMs}  
        &TokenFlow~\cite{geyer2024tokenflow}  &38.88 & 19.08 & 246.65 & 138.69 & 72.99 & 26.30 & 19.70 &\underline{4.21} &\underline{88.19} \\
        &DMT~\cite{yatim2024space}  &91.04 & 14.26 & 413.98 & 412.38 & 50.37 & \underline{26.86} & \textbf{20.38} &5.22 &80.23  \\
        &VidToMe~\cite{li2024vidtome}   &27.95 & 20.80 & 264.37 & 95.43 & 70.98 & \textbf{26.95} & 19.94 &4.82 &\textbf{88.97}  \\
        & AnyV2V~\cite{ku2024anyv2v}  &70.14 & 16.01 & 350.89 & 325.25 & 49.66 & 23.87 & 18.46 &4.62 &60.44  \\
        &VideoGrain~\cite{yang2025videograin} &\textbf{11.24} &\textbf{27.23} & \underline{180.61} & \textbf{26.13} & \underline{79.56} & 25.46 & 19.35 &\textbf{4.12} &87.38 \\
        \hline
        \multirow{2}{*}{\makecell{RFs \\ (Ours)}} 
        &Pyramid-Edit  &30.00 & 20.65 & 279.11 & 101.41 & 71.74 & \underline{26.86} & 18.97  &5.52 &79.55 \\
        &Wan-Edit      &\underline{14.33} & \underline{24.54} & \textbf{96.53} & \underline{40.36} & \textbf{82.33} & 26.85 &\underline{19.98}  &6.63 &87.94 \\
        \Xhline{1pt}
    \end{tabular}}
\end{table*}

\begin{table*}[h]
    \centering
    \begin{minipage}{0.49\textwidth}
        \caption{{\color{BrickRed} Edit1}: Comparison of diffusion- and flow-based video editing methods for {\color{BrickRed} object replacement without non-rigid transformations} on the FiVE benchmark using FiVE-Acc metrics.}
        \label{tab:baselines-edit1-five-acc}
        \vspace{-2mm}
        \footnotesize
        \renewcommand{\arraystretch}{1.2}
        \setlength{\tabcolsep}{3pt} 
        \begin{tabular}{l|cccc|c}
            \Xhline{1pt}
            \multicolumn{1}{l|}{Method} &{FiVE-YN} &{FiVE-MC} &{FiVE-$\cup$} &{FiVE-$\cap$} & FiVE-Acc$\uparrow$ \\
            \Xhline{1pt}
            \hline
            TokenFlow~\cite{geyer2024tokenflow}  &30.30 & 48.48 & 49.49 & 29.29 & 39.39 \\
            DMT~\cite{yatim2024space}  &\textbf{55.95} & \textbf{79.76} & \textbf{79.76} & \textbf{55.95} & \textbf{67.86} \\
            VidToMe~\cite{li2024vidtome}   &25.25 & 47.47 & 47.47 & 25.25 & 36.36 \\
            AnyV2V~\cite{ku2024anyv2v}  &27.27 & 42.42 & 44.44 & 25.25 & 34.85 \\
            VideoGrain~\cite{yang2025videograin} &40.00 & 62.00 & 62.00 & 40.00 & 51.00 \\
            \hline
            Pyramid-Edit  &27.27 & 53.54 & 55.56 & 25.25 & 40.40 \\
            Wan-Edit      &\underline{41.41} & \underline{62.63} & \underline{62.63} & \underline{41.41} & \underline{52.02} \\
            \Xhline{1pt}
        \end{tabular}
    \end{minipage}
    \hfill
    \begin{minipage}{0.49\textwidth}
        \centering
        \caption{{\color{orange} Edit2}: Comparison of diffusion- and flow-based video editing methods for {\color{orange} object replacement with non-rigid transformations} on the FiVE benchmark using FiVE-Acc metrics.}
        \label{tab:baselines-edit2-five-acc}
        \vspace{-2mm}
        \footnotesize
        \renewcommand{\arraystretch}{1.2}
        \setlength{\tabcolsep}{3pt} 
        \begin{tabular}{l|cccc|c}
            \Xhline{1pt}
            \multicolumn{1}{l|}{Method} &{FiVE-YN} &{FiVE-MC} &{FiVE-$\cup$} &{FiVE-$\cap$} & FiVE-Acc$\uparrow$ \\
            \Xhline{1pt}
            \hline
            TokenFlow~\cite{geyer2024tokenflow}  &18.18 & 37.37 & 38.38 & 17.17 & 27.78 \\
            DMT~\cite{yatim2024space}  &\textbf{41.49} & \underline{65.96} & \underline{68.09} & \textbf{39.36} & \textbf{53.72} \\
            VidToMe~\cite{li2024vidtome}   &23.23 & 41.41 & 43.43 & 21.21 & 32.32 \\
            AnyV2V~\cite{ku2024anyv2v}  &13.13 & 26.26 & 30.30 & 9.09 & 19.70 \\
            VideoGrain~\cite{yang2025videograin} &12.24 & 26.53 & 26.53 & 12.24 & 19.39 \\
            \hline
            Pyramid-Edit  &30.30 & 60.61 & 60.61 & 30.30 & 45.45 \\
            Wan-Edit      &\underline{36.36} & \textbf{67.68} & \textbf{68.69} & \underline{35.35} & \underline{52.02} \\
            \Xhline{1pt}
        \end{tabular}
    \end{minipage}
\end{table*}
\begin{table*}[h]
    \centering
    \caption{{\color{cyan} Edit3}: Comparison of diffusion- and flow-based video editing methods for {\color{cyan} object color changes} on the FiVE benchmark.}
    \resizebox{0.99\textwidth}{!}{
    \renewcommand{\arraystretch}{1.2}
    \setlength{\tabcolsep}{2.8pt} 
    \begin{tabular}{cl|c|cccc|cc|c|cc}
        \Xhline{1pt}
        \multicolumn{2}{c|}{\multirow{2}{*}{\textbf{Methods}}}  & \textbf{{Structure}} &\multicolumn{4}{c|}{\textbf{Background Preservation}} &\multicolumn{2}{c|}{\textbf{Text Alignment}} &\multicolumn{1}{c|}{\textbf{IQA}} &\multicolumn{1}{c}{\textbf{Temp. Consis.}} \\
        & &Dist.$_{\times 10^3}$$\downarrow$ &PSNR$\uparrow$ &LPIPS$_{\times 10^3}$$\downarrow$ &MSE$_{\times 10^4}$$\downarrow$ &SSIM$_{\times 10^2}$$\uparrow$ &CLIPS.$\uparrow$ &CLIPS.$_{\text{edit}}\uparrow$ &NIQE$\downarrow$ & Motion Fidelity S.$_{\times 10^2}$$\uparrow$  \\
        \Xhline{1pt} 
        &Source Videos &0.00 & $\infty$ &0.00 &0.00 &100.00 &26.12 & 20.64 &6.33 &93.76 \\
        \Xhline{1pt} 
        \multirow{5}{*}{DMs} 
        &TokenFlow~\cite{geyer2024tokenflow}  &35.03 & 19.07 & 262.05 & 138.21 & 72.71 & 27.72 & 21.68 &\textbf{4.02} &88.37 \\
        &DMT~\cite{yatim2024space}  &85.75 & 14.23 & 413.08 & 413.91 & 50.68 & 28.18 & 22.10  &5.20 &81.71 \\
        &VidToMe~\cite{li2024vidtome}   &21.90 & 21.17 & 261.94 & 89.81 & 72.20 & \textbf{28.45} & \underline{22.16}  &4.68 &\textbf{89.26} \\
        & AnyV2V~\cite{ku2024anyv2v}  & 79.16 & 14.37 & 411.83 & 455.53 & 46.82 & 26.58 & 20.59  &4.65 &61.13 \\
        &VideoGrain~\cite{yang2025videograin} &\underline{14.20} & \textbf{27.08} & \underline{185.38} & \textbf{25.95} & \underline{79.37} & \underline{28.44} & \textbf{22.23} &\underline{4.09} &87.41 \\
        \hline
        \multirow{2}{*}{\makecell{RFs \\ (Ours)}} 
        &Pyramid-Edit  &29.37 & 20.85 & 278.17 & 96.16 & 72.11 & 28.10 & 21.13  &5.49 &78.80 \\
        &Wan-Edit      &\textbf{11.63} & \underline{25.32} & \textbf{90.82} & \underline{35.77} & \textbf{83.04} & 27.39 & 22.04  &6.58 &\underline{88.59} \\
        \Xhline{1pt}
    \end{tabular}}

    \label{tab:baselines-edit3}
\end{table*}

\begin{table*}[h]
    \centering
    \caption{{\color{Yellow} Edit4}: Comparison of diffusion- and flow-based video editing methods for {\color{Yellow} object material changes} on the FiVE benchmark.}
    \resizebox{0.99\textwidth}{!}{
    \renewcommand{\arraystretch}{1.2}
    \setlength{\tabcolsep}{2.8pt} 
    \begin{tabular}{cl|c|cccc|cc|c|cc}
        \Xhline{1pt}
        \multicolumn{2}{c|}{\multirow{2}{*}{\textbf{Methods}}}  & \textbf{{Structure}} &\multicolumn{4}{c|}{\textbf{Background Preservation}} &\multicolumn{2}{c|}{\textbf{Text Alignment}} &\multicolumn{1}{c|}{\textbf{IQA}} &\multicolumn{1}{c}{\textbf{Temp. Consis.}} \\
        & &Dist.$_{\times 10^3}$$\downarrow$ &PSNR$\uparrow$ &LPIPS$_{\times 10^3}$$\downarrow$ &MSE$_{\times 10^4}$$\downarrow$ &SSIM$_{\times 10^2}$$\uparrow$ &CLIPS.$\uparrow$ &CLIPS.$_{\text{edit}}\uparrow$ &NIQE$\downarrow$ & Motion Fidelity S.$_{\times 10^2}$$\uparrow$  \\
        \Xhline{1pt} 
        &Source Videos &0.00 & $\infty$ &0.00 &0.00 &100.00 &26.89 & 21.85 &6.33 &93.76 \\
        \Xhline{1pt}
        \multirow{5}{*}{DMs}
        &TokenFlow~\cite{geyer2024tokenflow}  &35.22 & 19.22 & 258.53 & 133.83 & 73.06 & 27.67 & 22.19 &\underline{4.10} &88.34 \\
        &DMT~\cite{yatim2024space}    &84.34 & 14.15 & 408.88 & 417.24 & 50.24 & 27.33 & \underline{22.28} &5.14 &80.47 \\
        &VidToMe~\cite{li2024vidtome} &23.19 & 20.71 & 273.64 & 98.70 & 69.49 & \underline{27.82} & 21.91&4.70 &\textbf{89.24} \\
        & AnyV2V~\cite{ku2024anyv2v}  &71.97 & 15.53 & 354.39 & 382.99 & 50.33 & 26.42 & 20.99&4.60 &62.14 \\
        &VideoGrain~\cite{yang2025videograin} &\textbf{10.34} & \textbf{27.21} & \underline{185.78} & \textbf{25.61} & \underline{79.13} & 27.44 & 21.15 &\textbf{4.02} &87.55 \\
        \hline
        \multirow{2}{*}{\makecell{RFs \\ (Ours)}} 
        &Pyramid-Edit  &28.39 & 20.74 & 277.73 & 98.54 & 72.04 & \textbf{28.07} & 21.48 &5.44 &78.27 \\
        &Wan-Edit      &\underline{10.66} & \underline{25.45} & \textbf{89.72} & \underline{34.19} & \textbf{83.31} & 27.55 & \textbf{22.35}  &6.57 &\underline{88.83} \\
        \Xhline{1pt}
    \end{tabular}}

    \label{tab:baselines-edit4}
\end{table*}

\begin{table*}[h]

    \begin{minipage}{0.49\textwidth}
        \centering
        \caption{{\color{cyan} Edit3}: Comparison of diffusion- and flow-based video editing methods for {\color{cyan} object color changes} on the FiVE benchmark using FiVE-Acc metrics.}
        \label{tab:baselines-edit3-five-acc}
        \vspace{-2mm}
        \footnotesize
        \renewcommand{\arraystretch}{1.2}
        \setlength{\tabcolsep}{3pt} 
        \begin{tabular}{l|cccc|c}
            \Xhline{1pt}
            \multicolumn{1}{l|}{Method} &{FiVE-YN} &{FiVE-MC} &{FiVE-$\cup$} &{FiVE-$\cap$} & FiVE-Acc$\uparrow$ \\
            \Xhline{1pt}
            \hline
            TokenFlow~\cite{geyer2024tokenflow}  &34.34 & 46.46 & 48.48 & 32.32 & 40.40 \\
            DMT~\cite{yatim2024space}  &55.06 & 61.80 & 64.04 & 52.81 & 58.43 \\
            VidToMe~\cite{li2024vidtome}   &36.36 & 42.42 & 43.43 & 35.35 & 39.39 \\
            AnyV2V~\cite{ku2024anyv2v}  &54.55 & \underline{64.65} & 67.68 & 51.52 & 59.60 \\
            VideoGrain~\cite{yang2025videograin} &\textbf{82.00} & \textbf{90.00} & \textbf{92.00} & \textbf{80.00} & \textbf{86.00} \\
            \hline
            Pyramid-Edit  &59.60 & 57.58 & 66.67 & 50.51 & 58.59 \\
            Wan-Edit      &\underline{62.63} & 63.64 & \underline{68.69} & \underline{57.58} & \underline{63.13} \\
            \Xhline{1pt}
        \end{tabular}
    \end{minipage}
    \hfill
    \begin{minipage}{0.49\textwidth}
        \centering
        \caption{{\color{Yellow} Edit4}: Comparison of diffusion- and flow-based video editing methods for {\color{Yellow} object material changes} on the FiVE benchmark using FiVE-Acc metrics.}
        \label{tab:baselines-edit4-five-acc}
        \vspace{-2mm}
        \footnotesize
        \renewcommand{\arraystretch}{1.2}
        \setlength{\tabcolsep}{3pt} 
        \begin{tabular}{l|cccc|c}
            \Xhline{1pt}
            \multicolumn{1}{l|}{Method} &{FiVE-YN} &{FiVE-MC} &{FiVE-$\cup$} &{FiVE-$\cap$} & FiVE-Acc$\uparrow$ \\
            \Xhline{1pt}
            \hline
            TokenFlow~\cite{geyer2024tokenflow}  &11.11 & 26.26 & 29.29 & 8.08 & 18.69 \\
            DMT~\cite{yatim2024space}  &11.76 & 47.06 & 48.24 & 10.59 & 29.41 \\
            VidToMe~\cite{li2024vidtome}   &13.13 & 36.36 & 38.38 & 11.11 & 24.75 \\
            AnyV2V~\cite{ku2024anyv2v}  &\underline{23.23} & \textbf{63.64} & \textbf{64.65} & \underline{22.22} & 43.43 \\
            VideoGrain~\cite{yang2025videograin} &\textbf{26.53} & 40.82 & 40.82 & \textbf{26.53} & 33.67 \\
            \hline
            Pyramid-Edit  &18.18 & \underline{54.55} & \underline{57.58} & 15.15 & \underline{36.36} \\
            Wan-Edit      &19.19 & \textbf{43.43} & 45.45 & 17.17 & 31.31 \\
            \Xhline{1pt}
        \end{tabular}
    \end{minipage}

\end{table*}
\begin{table*}[h]
    \centering
    \caption{{\color{Green} Edit5}: Comparison of diffusion- and flow-based video editing methods for {\color{Green} object addition} on the FiVE benchmark.}
    \resizebox{0.99\textwidth}{!}{
    \renewcommand{\arraystretch}{1.2}
    \setlength{\tabcolsep}{2.8pt} 
    \begin{tabular}{cl|c|cccc|cc|c|cc}
        \Xhline{1pt}
        \multicolumn{2}{c|}{\multirow{2}{*}{\textbf{Methods}}}  & \textbf{{Structure}} &\multicolumn{4}{c|}{\textbf{Background Preservation}} &\multicolumn{2}{c|}{\textbf{Text Alignment}} &\multicolumn{1}{c|}{\textbf{IQA}} &\multicolumn{1}{c}{\textbf{Temp. Consis.}} \\
        & &Dist.$_{\times 10^3}$$\downarrow$ &PSNR$\uparrow$ &LPIPS$_{\times 10^3}$$\downarrow$ &MSE$_{\times 10^4}$$\downarrow$ &SSIM$_{\times 10^2}$$\uparrow$ &CLIPS.$\uparrow$ &CLIPS.$_{\text{edit}}\uparrow$ &NIQE$\downarrow$ & Motion Fidelity S.$_{\times 10^2}$$\uparrow$  \\
        \Xhline{1pt} 
        &Source Videos &0.00 & $\infty$ &0.00 &0.00 &100.00 &23.52 &19.58 &5.58 &97.86 \\
        \Xhline{1pt}
        \multirow{5}{*}{DMs}
        &TokenFlow~\cite{geyer2024tokenflow}  &36.76 & 18.55 & 295.59 & 152.12 & 67.04 & \textbf{25.32} & \underline{21.07} &\textbf{3.25} &96.71 \\ 
        &DMT~\cite{yatim2024space}  &93.29 & 15.82 & 413.62 & 275.22 & 47.90 & \underline{25.14} & 20.85 &5.14 &91.70 \\
        &VidToMe~\cite{li2024vidtome}   &\underline{22.31} & 20.55 & 275.19 & \underline{93.10} & 64.19 & 24.38 & 19.73 &4.31 &97.13 \\
        &AnyV2V~\cite{ku2024anyv2v}  &55.00 & 16.68 & 328.01 & 249.76 & 48.49 & 25.01 & 20.40 &\underline{4.10} &62.76 \\
        &VideoGrain~\cite{yang2025videograin} &\textbf{18.03} & \textbf{26.42} & \underline{208.18} & \textbf{27.36} & \textbf{74.44} & 22.30 & 18.52 &\underline{4.10} &\textbf{98.36} \\
        \hline
        \multirow{2}{*}{\makecell{RFs \\ (Ours)}} 
        &Pyramid-Edit  &29.24 & 20.33 & 293.97 & 100.99 & 65.23 & 24.83 & 20.25 &5.39 &89.90 \\
        &Wan-Edit      &23.04 & \underline{20.70} & \textbf{139.79} & 93.43 & \underline{73.55} & 25.09 & \textbf{21.32} &5.84 &\underline{97.38}\\
        \Xhline{1pt}
    \end{tabular}}

    \label{tab:baselines-edit5}
\end{table*}

\begin{table*}[h]
    \centering
    \caption{{\color{Purple} Edit6}: Comparison of diffusion- and flow-based video editing methods for {\color{Purple} object removal} on the FiVE benchmark.}
    \resizebox{0.99\textwidth}{!}{
    \renewcommand{\arraystretch}{1.2}
    \setlength{\tabcolsep}{2.8pt} 
    \begin{tabular}{cl|c|cccc|cc|c|cc}
        \Xhline{1pt}
        \multicolumn{2}{c|}{\multirow{2}{*}{\textbf{Methods}}}  & \textbf{{Structure}} &\multicolumn{4}{c|}{\textbf{Background Preservation}} &\multicolumn{2}{c|}{\textbf{Text Alignment}} &\multicolumn{1}{c|}{\textbf{IQA}} &\multicolumn{1}{c}{\textbf{Temp. Consis.}} \\
        & &Dist.$_{\times 10^3}$$\downarrow$ &PSNR$\uparrow$ &LPIPS$_{\times 10^3}$$\downarrow$ &MSE$_{\times 10^4}$$\downarrow$ &SSIM$_{\times 10^2}$$\uparrow$ &CLIPS.$\uparrow$ &CLIPS.$_{\text{edit}}\uparrow$ &NIQE$\downarrow$ & Motion Fidelity S.$_{\times 10^2}$$\uparrow$  \\
        \Xhline{1pt} 
        &Source Videos &0.00 & $\infty$ &0.00 &0.00 &100.00 &24.91 & 21.06 &6.79 &92.63 \\
        \Xhline{1pt}
        \multirow{5}{*}{DMs} 
        &TokenFlow~\cite{geyer2024tokenflow}  &31.61 & 19.33 & 261.55 & 131.90 & 76.63 & 24.70 & 21.02 &\underline{4.42} &84.15 \\
        &DMT~\cite{yatim2024space}  &75.91 & 15.43 & 367.21 & 315.51 & 59.05 & 25.22 & \textbf{21.53} &5.49 &77.76 \\
        &VidToMe~\cite{li2024vidtome}   &15.03 & 22.35 & 247.12 & 66.99 & 75.80 & \textbf{25.67} &\underline{21.34} &4.91 &\textbf{85.12} \\
        &AnyV2V~\cite{ku2024anyv2v}  &80.71 & 16.97 & 300.65 & 293.93 & 58.38 & 22.06 & 17.93 &4.93 &54.07 \\
        &VideoGrain~\cite{yang2025videograin} &\underline{8.86} & \underline{27.46} & \underline{167.21} & \underline{18.76} & \underline{83.23} & 23.05 & 19.22 &\textbf{4.06} &82.46 \\
        \hline
        \multirow{2}{*}{\makecell{RFs \\ (Ours)}} 
        &Pyramid-Edit  &26.63 & 21.58 & 254.41 & 80.54 & 76.65 & \underline{25.60} & 19.26  &5.55 &75.52 \\
        &Wan-Edit      &\textbf{2.02} & \textbf{32.62} & \textbf{57.12} & \textbf{7.63} & \textbf{90.52} & 24.29 & 20.31 &7.02 &\underline{84.50} \\
        \Xhline{1pt}
    \end{tabular}}

    \label{tab:baselines-edit6}
\end{table*}

\begin{table*}[h]
    \begin{minipage}{0.49\textwidth}
        \centering
        \caption{{\color{Green} Edit5}: Comparison of diffusion- and flow-based video editing methods for {\color{Green} object addition} on the FiVE benchmark using FiVE-Acc metrics.}
        \label{tab:baselines_edit5_five_acc}
        \vspace{-2mm}
        \footnotesize
        \renewcommand{\arraystretch}{1.2}
        \setlength{\tabcolsep}{3pt} 
        \begin{tabular}{l|cccc|c}
            \Xhline{1pt}
            \multicolumn{1}{l|}{Method} &{FiVE-YN} &{FiVE-MC} &{FiVE-$\cup$} &{FiVE-$\cap$} & FiVE-Acc$\uparrow$ \\
            \Xhline{1pt}
            \hline
            TokenFlow~\cite{geyer2024tokenflow}  &22.22 & 44.44 & 44.44 & 22.22 & 33.33 \\
            DMT~\cite{yatim2024space}      &44.44 & \textbf{77.78} & \underline{77.78} & 44.44 & 61.11 \\
            VidToMe~\cite{li2024vidtome}   &22.22 & 33.33 & 44.44 & 11.11 & 27.78 \\
            AnyV2V~\cite{ku2024anyv2v}     &55.56 & \underline{55.56} & 66.67 & 44.44 & 55.56\\
            VideoGrain~\cite{yang2025videograin} &22.22 & 33.33 & 33.33 & 22.22 & 27.78 \\
            \hline
            Pyramid-Edit  &\underline{66.67} & \textbf{77.78} & \underline{77.78} & \underline{66.67} & \underline{72.22} \\
            Wan-Edit      &\textbf{88.89} & \textbf{77.78} & \textbf{88.89} & \textbf{77.78} & \textbf{83.33} \\
            \Xhline{1pt}
        \end{tabular}
    \end{minipage}
    \hfill
    \begin{minipage}{0.49\textwidth}
        \centering
        \caption{{\color{Purple} Edit6}: Comparison of diffusion- and flow-based video editing methods {\color{Purple} object removal} on the FiVE benchmark using FiVE-Acc metrics.}
        \label{tab:baselines-edit6-five-acc}
        \vspace{-2mm}
        \footnotesize
        \renewcommand{\arraystretch}{1.2}
        \setlength{\tabcolsep}{3pt} 
        \begin{tabular}{l|cccc|c}
            \Xhline{1pt}
            \multicolumn{1}{l|}{Method} &{FiVE-YN} &{FiVE-MC} &{FiVE-$\cup$} &{FiVE-$\cap$} & FiVE-Acc$\uparrow$ \\
            \Xhline{1pt}
            \hline
            TokenFlow~\cite{geyer2024tokenflow}  &0.00 & 10.00 & 10.00 & 0.00 & 5.00 \\
            DMT~\cite{yatim2024space}  &0.00 & \textbf{40.00} & \textbf{40.00} & 0.00 & \textbf{20.00} \\
            VidToMe~\cite{li2024vidtome}  &0.00 & 0.00 & 0.00 & 0.00 & 0.00 \\
            AnyV2V~\cite{ku2024anyv2v}  &\textbf{10.00} & \underline{20.00} & \underline{20.00} & \textbf{10.00} & \underline{15.00} \\
            VideoGrain~\cite{yang2025videograin} &0.00 & 11.11 & 11.11 & 0.00 & 5.56 \\
            \hline
            Pyramid-Edit  &0.00 & \underline{20.00} & \underline{20.00} & 0.00 & 10.00 \\
            Wan-Edit      &0.00 & 0.00 & 0.00 & 0.00 & 0.00 \\
            \Xhline{1pt}
        \end{tabular}
    \end{minipage}

\end{table*}

In this section, we present additional experimental results and provide a comprehensive analysis of the performance of all baseline methods on our proposed FiVE benchmark. For clarity, we define six editing types, referred to as Edit 1–6: rigid transformation (e.g., car to bus), non-rigid transformation (e.g., car to elephant), color change (e.g., black to red), attribute change (e.g., car to a wooden texture), object addition, and object removal. We conduct a comparative analysis of various video editing methods across these categories using multiple metrics. Additionally, we evaluate diffusion- and flow-based approaches on the proposed FiVE benchmark and FiVE-Acc metric.

Tables~\ref{tab:baselines-edit1} and \ref{tab:baselines-edit1-five-acc} present the results for {\color{BrickRed} Edit1}: Object replacement without non-rigid transformations (e.g., replacing a car with a bus). Our proposed Wan-Edit achieves background preservation and text alignment comparable to VideoGrain, while exhibiting superior motion fidelity. Since Wan-Edit better retains the original video background, its IQA scores remain consistent with those of the source video. Regarding the FiVE-Acc metrics, which assess the accuracy of successful edits, the most competitive method is DMT, which optimizes editing based on the input text. The training-free Wan-Edit and VideoGrain achieve similar results, slightly trailing DMT but significantly outperforming other methods. Overall, DMT offers the best text-vision alignment but requires optimization, whereas VideoGrain and Wan-Edit strike a strong balance across various fine-grained video editing metrics. Notably, Wan-Edit stands out for its superior efficiency, delivering faster and more stable results compared to VideoGrain.

Tables~\ref{tab:baselines-edit2} and \ref{tab:baselines-edit2-five-acc} present the results of all compared methods on {\color{orange} Edit2}: Object Replacement with Non-Rigid Transformations, revealing similar conclusions to {\color{BrickRed} Edit1}: Object Replacement with Rigid Transformations. However, {\color{orange} Edit2} is more challenging than {\color{BrickRed} Edit1} due to the complexity of non-rigid transformations. This increased difficulty results in a noticeable drop in text-vision alignment, motion fidelity scores, and the editing success rate (FiVE-Acc) metrics compared to {\color{BrickRed} Edit1}.  In terms of the editing success rate (FiVE-Acc), DMT and Wan-Edit achieve 67.86\% and 52.02\%, respectively, on {\color{BrickRed} Edit1}, with DMT outperforming Wan-Edit by approximately 15\%. However, on {\color{orange} Edit2}, DMT drops significantly to 53.72\%, while Wan-Edit remains stable. This indicates Wan-Edit's robustness in handling non-rigid transformations, maintaining consistent performance even in more challenging editing scenarios.

Similarly, consistent trends are observed across other editing types, {\color{cyan} Edit3} (color changes) and {\color{Yellow} Edit4} (object material changes) as shown in Tables \ref{tab:baselines-edit3} - \ref{tab:baselines-edit4-five-acc}, further reinforcing our conclusions. Regarding the FiVE-Acc metric, color changes ({\color{cyan} Edit3}) achieve the highest editing success rate among all types, with VideoGrain reaching 86\% and Wan-Edit at 63\%. In contrast, object material changes ({\color{Yellow} Edit4}) are significantly more challenging, with AnyV2V achieving the highest success rate at 43\%, followed by Pyramid-Edit at 36\%. This difficulty arises because object material changes often require modifying mid- and low-frequency noise, making training-free methods highly sensitive to parameters, which leads to lower success rates.

Tables \ref{tab:baselines-edit5} - \ref{tab:baselines-edit6-five-acc} compare the results of {\color{Green} Edit5} (object addition) and {\color{Purple} Edit6} (object removal). For object addition ({\color{Green} Edit5}), VideoGrain and Wan-Edit achieve the highest scores in background preservation and motion fidelity, while TokenFlow performs best in text-vision alignment and IQA. In terms of FiVE-Acc, the RF-based methods Pyramid-Edit and Wan-Edit achieve success rates of 83\% and 72\%, respectively, whereas the highest-performing diffusion-based method, DMT, reaches only 61\%, highlighting the advantage of RF-based approaches in this task. For {\color{Purple} Edit6} (object removal), nearly all methods perform the worst among all editing types, with FiVE-Acc scores dropping below 20\%, indicating that object removal is one of the most challenging fine-grained editing tasks. This difficulty arises because removing an object requires precisely inpainting the occluded background while maintaining temporal coherence, which is particularly challenging for existing editing models. The visualizations in Fig. 4 of the main paper further confirm these findings.

In conclusion, our analysis ranks the difficulty of fine-grained video editing tasks, with color changes ({\color{cyan} Edit3}) being the easiest and object removal ({\color{Purple} Edit6}) the most challenging. Rigid object replacement ({\color{BrickRed} Edit1}) and object addition ({\color{Green} Edit5}) are relatively simple, while non-rigid transformations ({\color{orange} Edit2}) and material changes ({\color{Yellow} Edit4}) pose moderate challenges. The particularly low success rate of object removal highlights its complexity, requiring precise inpainting and temporal consistency.

\section{More Qualitative Results and Analysis}

\begin{figure*}[h]
    \centering
    \includegraphics[width=0.95\linewidth]{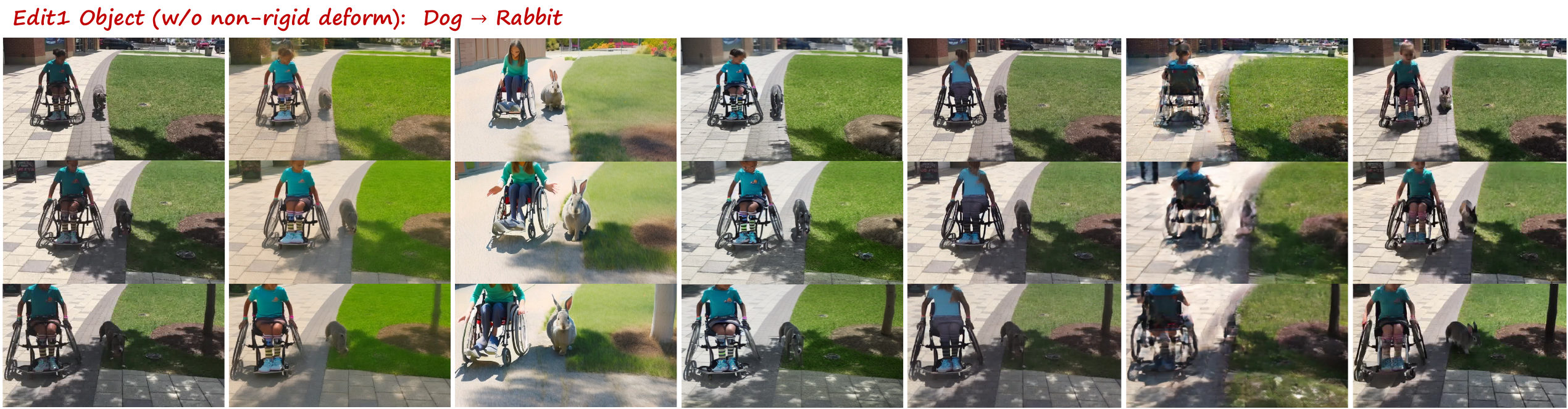}
    \includegraphics[width=0.95\linewidth]{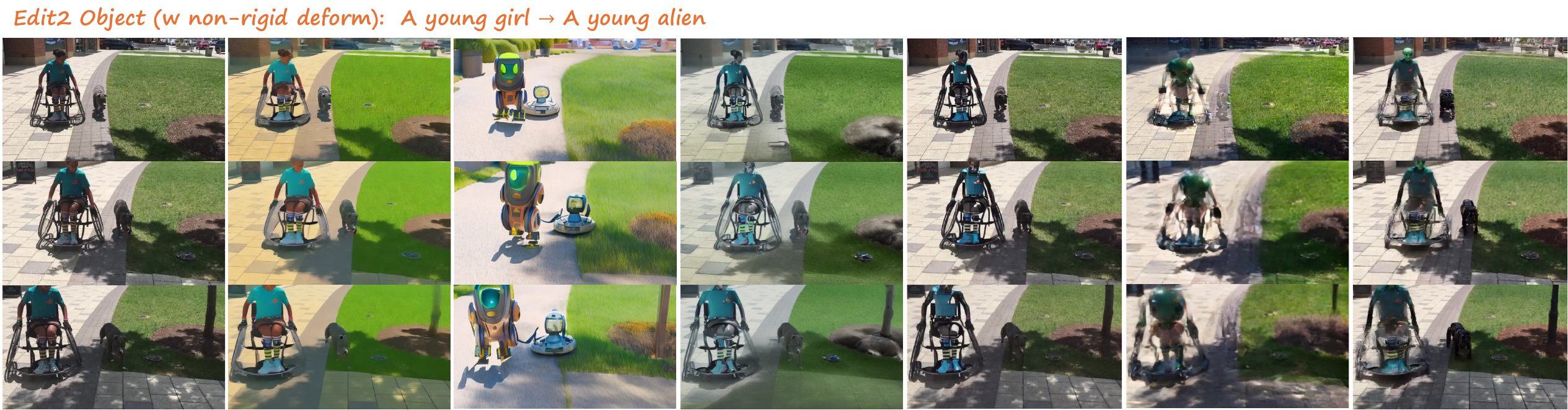}
    \includegraphics[width=0.95\linewidth]{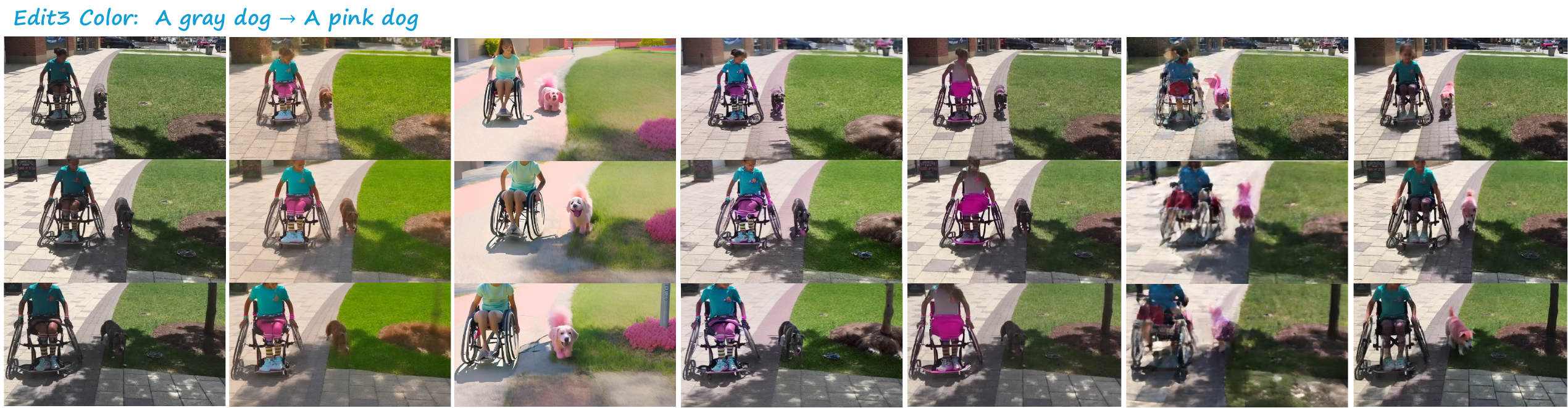}
    \includegraphics[width=0.95\linewidth]{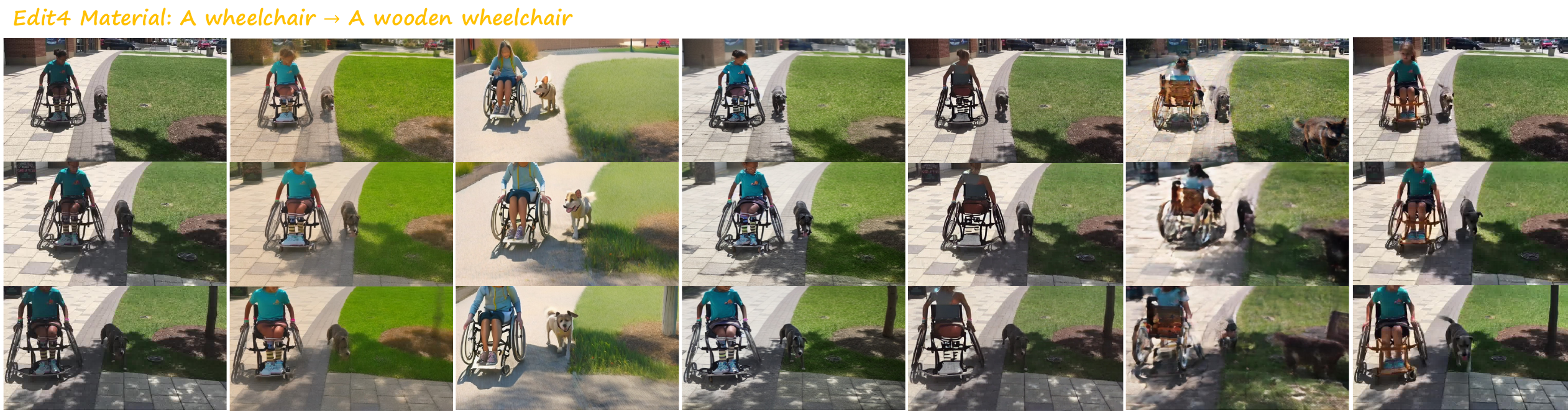}
    \includegraphics[width=0.95\linewidth]{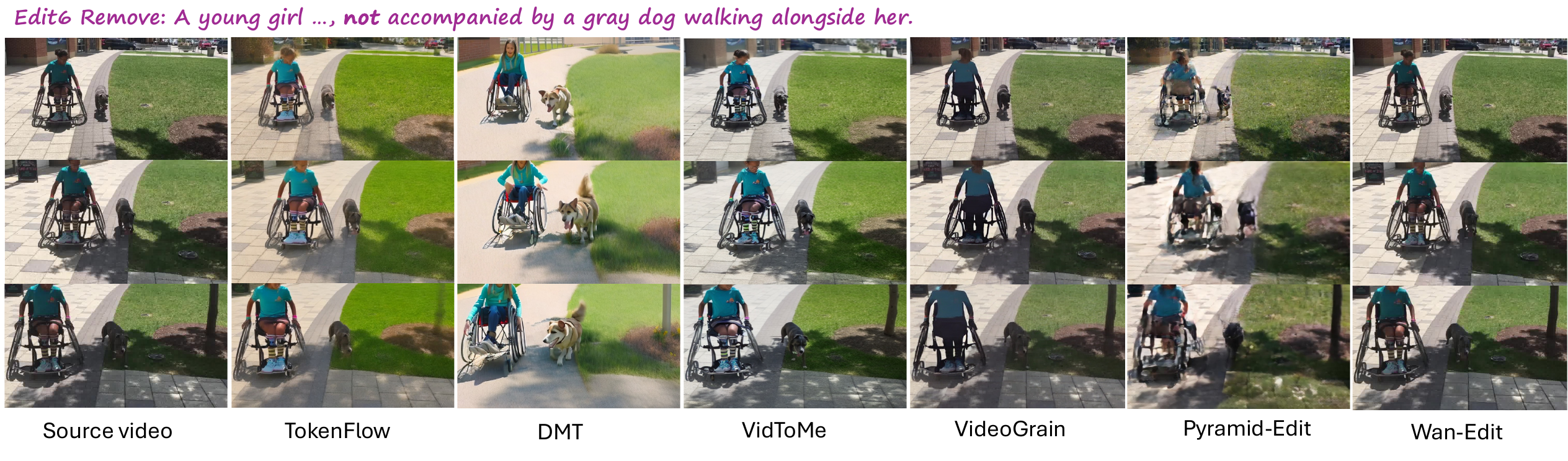}
    \vfill
    \vspace{-4mm}
    \caption{Editing results across five editing types and six high-performance comparison methods.}
    \label{fig:vis_results_02}
\end{figure*}

\begin{figure*}[h]
    \centering
    \includegraphics[width=0.95\linewidth]{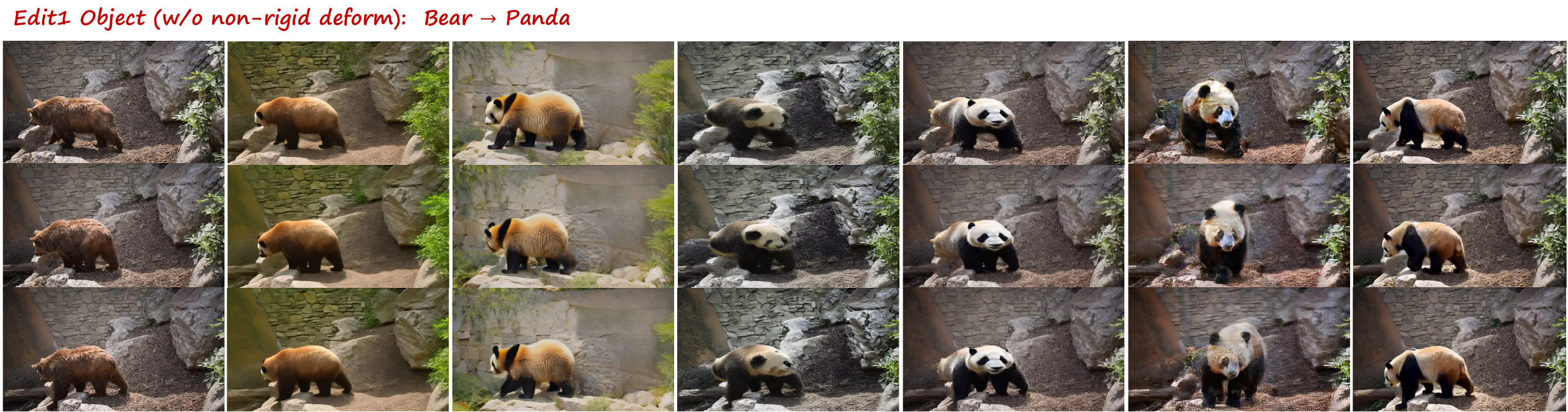}
    \includegraphics[width=0.95\linewidth]{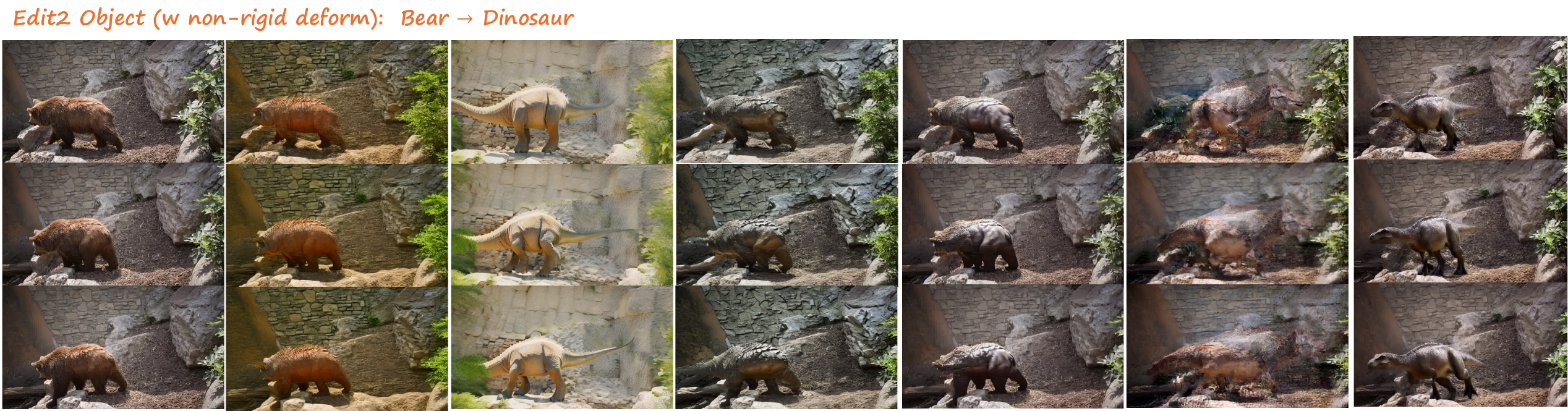}
    \includegraphics[width=0.95\linewidth]{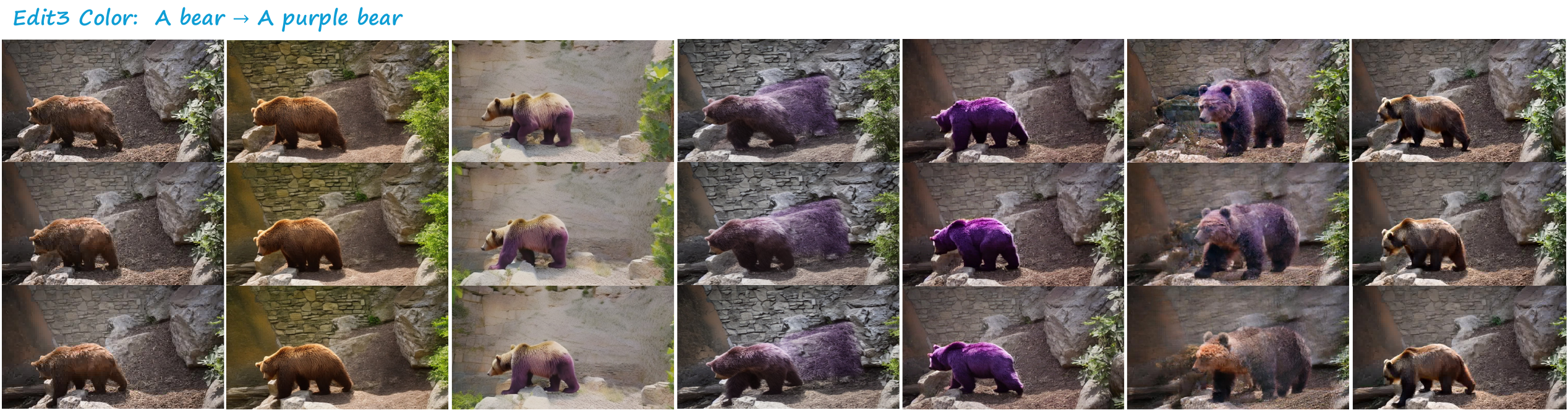}
    \includegraphics[width=0.95\linewidth]{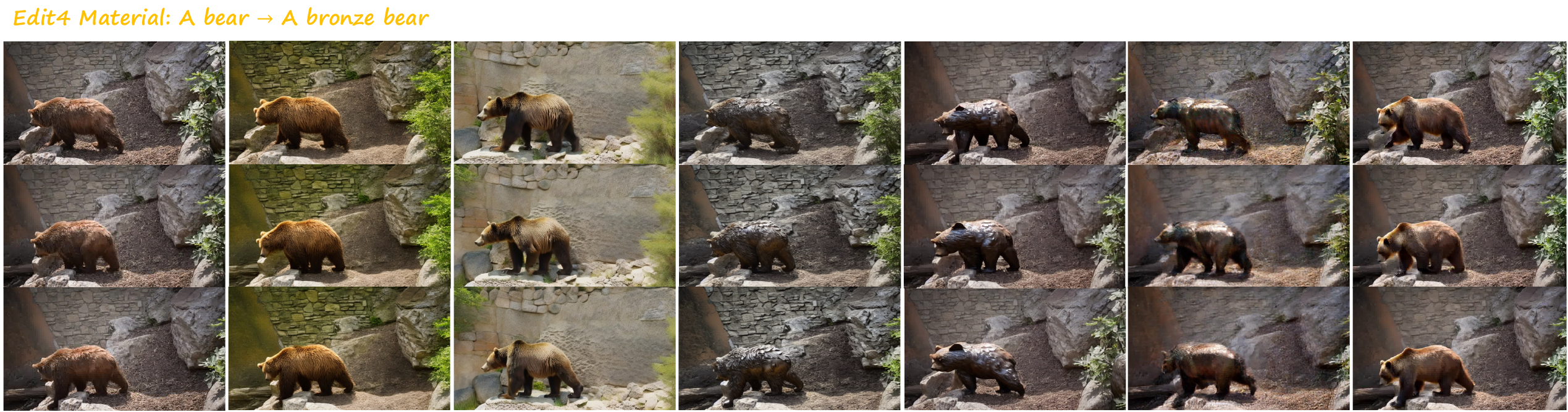}
    \includegraphics[width=0.95\linewidth]{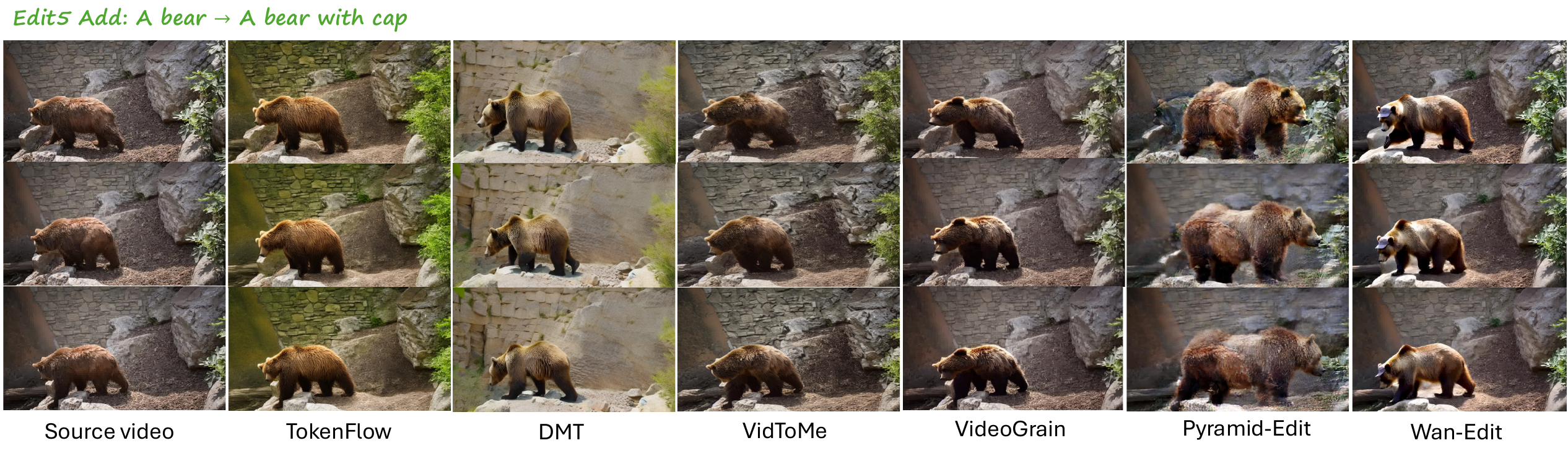}
    \vfill
    \vspace{-4mm}
    \caption{Editing results across five editing types and six high-performance comparison methods. {Wan-Edit} is the only method that succeeds in the object addition editing type. }
    \label{fig:vis_results_09}
\end{figure*}

\begin{figure*}[h]
    \centering
    \includegraphics[width=0.95\linewidth]{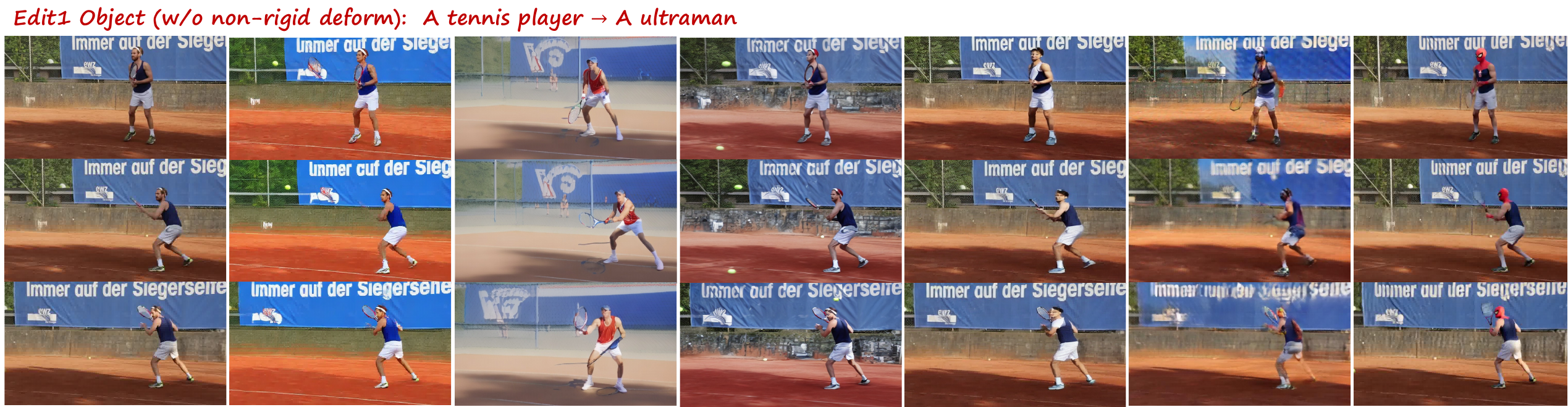}
    \includegraphics[width=0.95\linewidth]{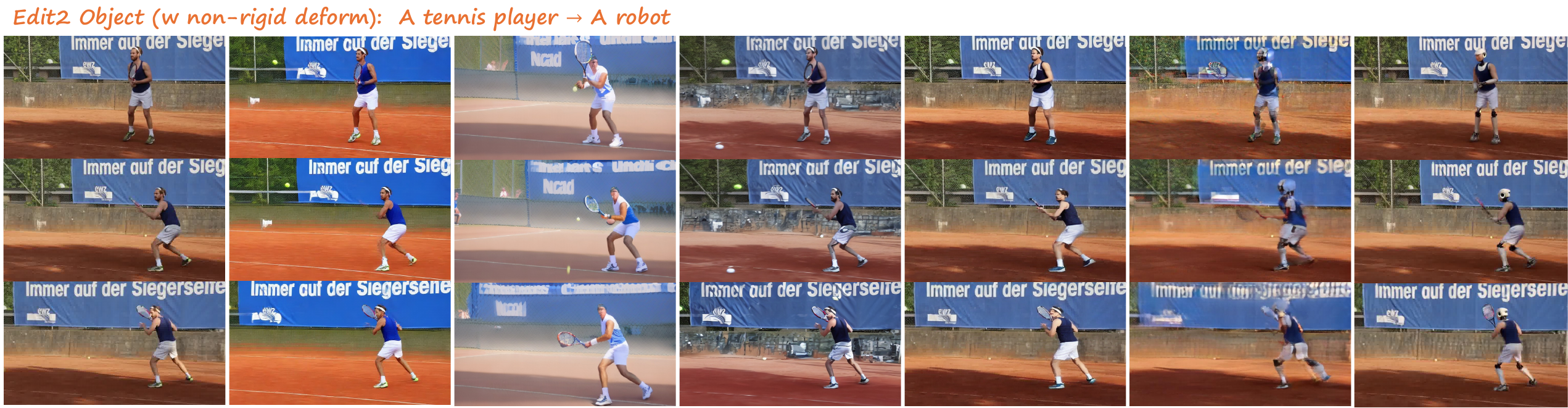}
    \includegraphics[width=0.95\linewidth]{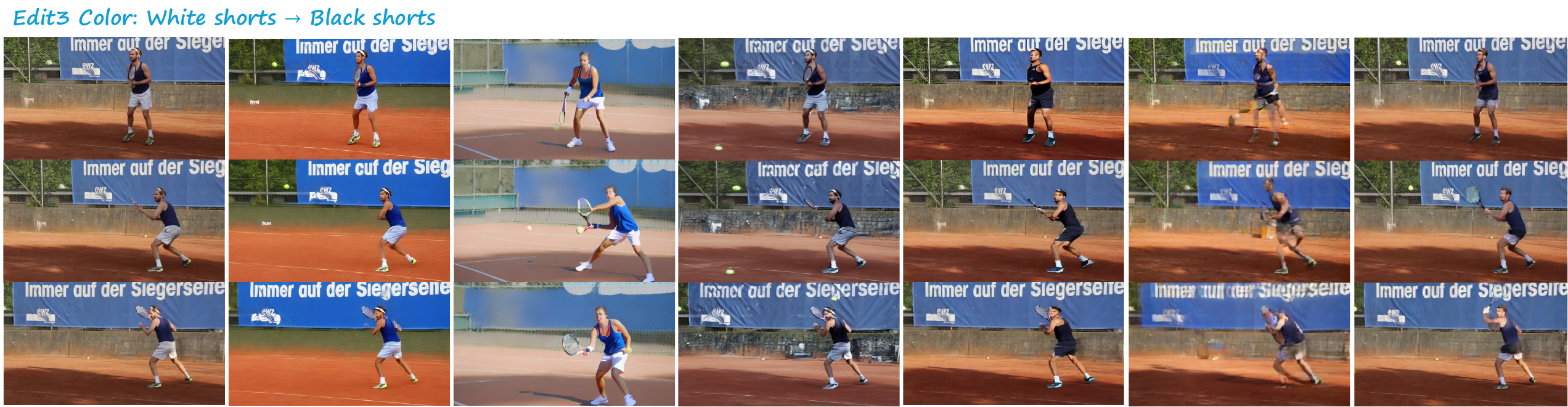}
    \includegraphics[width=0.95\linewidth]{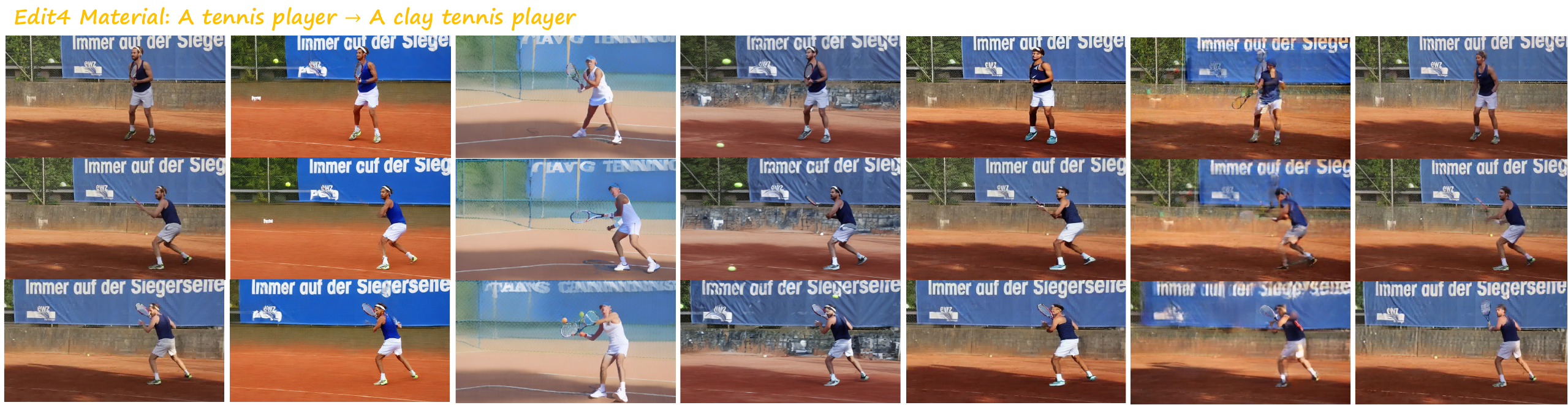}
    \includegraphics[width=0.95\linewidth]{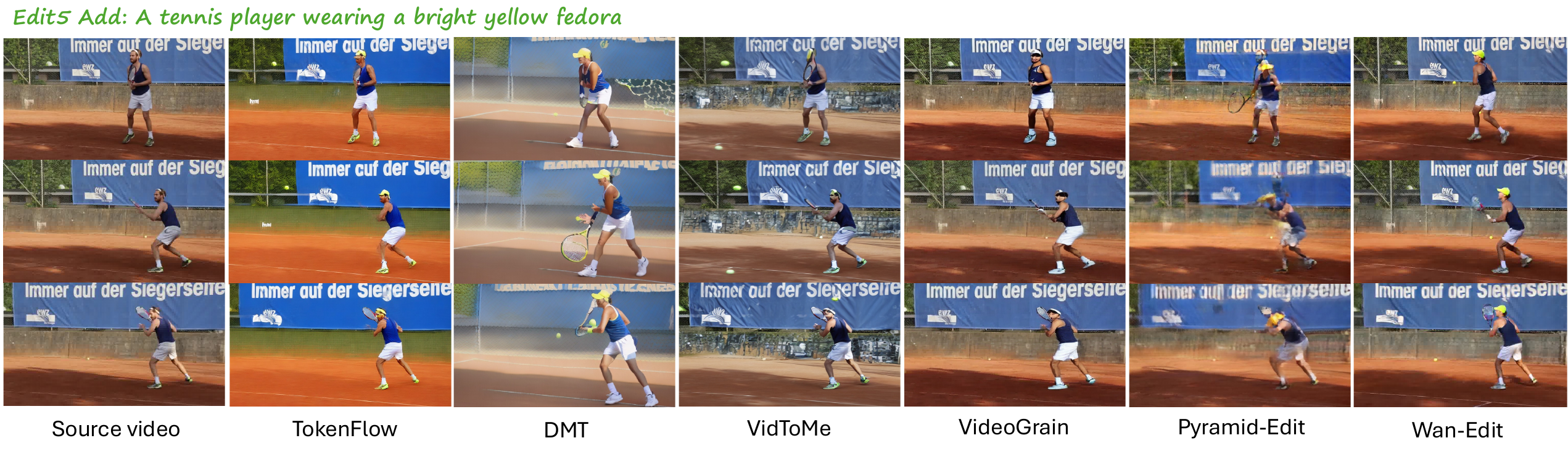}
    \vfill
    \vspace{-4mm}
    \caption{Editing results across five editing types and six high-performance comparison methods.}
    \label{fig:vis_results_19}
\end{figure*}

\begin{figure*}[h]
    \centering
    \includegraphics[width=\linewidth]{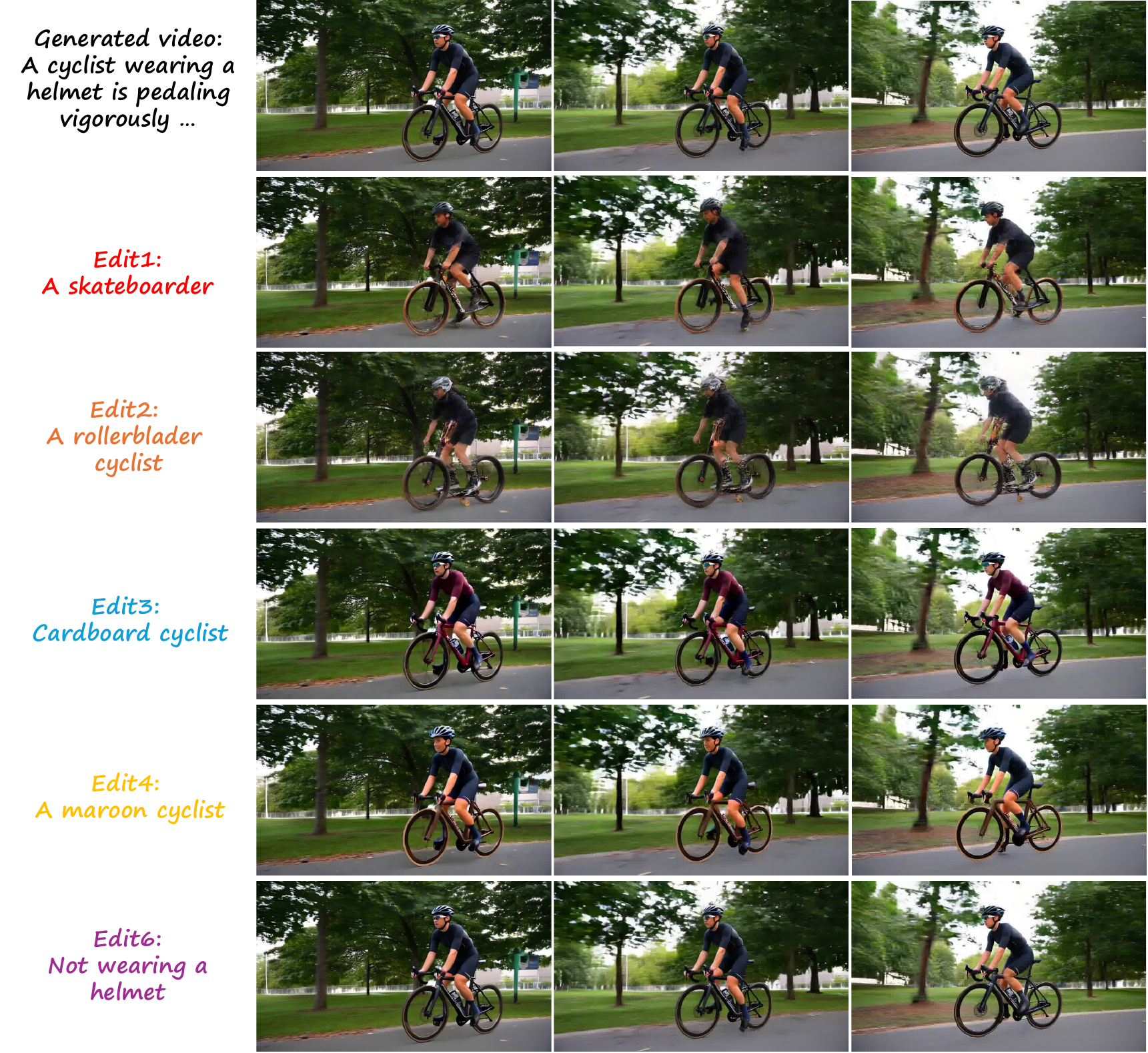}
    \caption{A generated video (first row) along with Wan-Edit's editing results across five editing types (rows 2-6).}
    \label{fig:vis_results_88}
\end{figure*}

We present the editing results across various editing types and comparison methods in Figs. \ref{fig:vis_results_02} - \ref{fig:vis_results_19} and Fig. \ref{fig:vis_results_88}. Figs. \ref{fig:vis_results_02} - \ref{fig:vis_results_19} compare the results of diffusion-based and RF-based editing methods across six editing types. These comparisons highlight the strengths and weaknesses of each method while also demonstrating the superiority of VideoGrain and our Wan-Edit. The videos shown in Fig. \ref{fig:vis_results_88} are generated examples and represent a particularly challenging editing case. Editing becomes difficult when the object occupies a significant portion of the video frame, as it requires modifying low-frequency noise while maintaining spatial and temporal consistency. This results in failure even for the best-performing Wan-Edit. To better showcase the dynamic consistency in video editing, more video demos are available on the anonymous website: \url{https://sites.google.com/view/five-benchmark}.

%


\end{document}